\pdfoutput=1

\documentclass[11pt]{article}

\usepackage[]{acl}

\usepackage{times}
\usepackage{latexsym}
\usepackage{amsfonts}
\usepackage{amssymb}
\usepackage{pdfpages}
\usepackage{tabularx}
\usepackage{amsmath,bm}

\usepackage[T1]{fontenc}

\usepackage[utf8]{inputenc}

\usepackage{microtype}
\usepackage{amsmath}
\usepackage{diagbox}
\usepackage{graphicx} 
\usepackage{cite}
\usepackage{booktabs}
\usepackage{multirow}
\usepackage{subcaption}
\usepackage{xurl}

\title{An Information-Theoretic Approach to Analyze NLP Classification Tasks}

\author{Luran Wang, Mark Gales, Vatsal Raina \\
  ALTA Institute, University of Cambridge \\
  \texttt{\{lw703,mjfg,vr311\}@cam.ac.uk} \\
  }

\begin{document}
\maketitle

\begin{abstract}

Understanding the importance of the inputs on the output is useful across many tasks. This work provides an information-theoretic framework to analyse the influence of inputs for text classification tasks.
Natural language processing (NLP) tasks take either a single element input or multiple element inputs to predict an output variable, where an element is a block of text. Each text element has two components: an associated semantic meaning and a linguistic realization.
Multiple-choice reading comprehension (MCRC) and sentiment classification (SC) are selected to showcase the framework.
For MCRC, it is found that the context influence on the output compared to the question influence reduces on more challenging datasets.
In particular, more challenging contexts allow a greater variation in complexity of questions.
Hence, test creators need to carefully consider the choice of the context when designing multiple-choice questions for assessment.
For SC, it is found the semantic meaning of the input text dominates (above 80\% for all datasets considered) compared to its linguistic realisation when determining the sentiment. 
The framework is made available at: \url{https://github.com/WangLuran/nlp-element-influence}.

\end{abstract}

\section{Introduction}

Natural Language Processing (NLP) requires machines to understand human language to
perform a specific task \citep{chowdhary2020natural}. NLP tasks typically take a single (such as summarization \citep{widyassari2022review}, sentiment classification \citep{wankhade2022survey}, machine translation \citep{stahlberg2020neural}) or multiple (such as reading comprehension \citep{baradaran2022survey}, question generation \citep{kurdi2020systematic}) text elements at the input and return a specific output. Each input text element can further be partitioned into its semantic content and the linguistic realization. The semantic meaning is the inherent meaning of the input element while the linguistic realization is the specific word choice to present the meaning in human language. Typically, there are several possible linguistic realizations for any semantic content. Therefore, for all NLP tasks, the output variable has contributions from at least two components: the semantic meaning of the element and the specific linguistic realization. Here, \textit{element} refers to a specific input text in an NLP task that is formed of exactly two \textit{component}s.

In this work, we analyze the relative sensitivity of the output variable to each of the input elements as well as in terms of the breakdown between the elemental semantic content and its corresponding linguistic realization. A theoretical information-theoretic approach is applied to find the shared information content between each input component and the output variable. Here, the information-theoretic approach is framed for NLP classification tasks where the set of input components influence the output probability distribution over a discrete set of classes. We select multiple-choice reading comprehension (MCRC) and sentiment classification as case studies for the analysis. 

MCRC requires the correct answer option to be selected based on several inputs element: the context paragraph, the question and the set of answer options. Multiple-choice (MC) assessments are a widely employed method for evaluating the competencies of candidates across diverse settings and tasks on a global scale \citep{Lai2017RACELR, richardson-etal-2013-mctest, sun2019dream, levesque2012winograd}. Given their consequential impact on real-world decisions, the selection of appropriate MC questions tailored to specific scenarios is important for content creators. Consequently, there is a need to comprehend the underlying factors that contribute to the complexity of these assessments. 

Complexity of an MC question is best modelled by the distribution over the answer options by human test takers. Therefore, by understanding the influence of each input element on the output distribution, content creators can be better informed to what extent the complexity of an MC question can be controlled from changing each of the input elements. Moreover, analyzing the contribution of the semantic content vs the linguistic realization on the output human distribution informs the impact of the specific word choice in the element on the question complexity.
However, it is not scalable to measure the variation in the output human distribution with variation in each of the input elements. 
\citet{liusie2023analysis} demonstrated that the output distribution of machine reading comprehension systems is aligned (with minimal re-shaping parameters) to the human distribution. Therefore, the information-theoretic framework is applied to understand the influence of each input element as well as the semantic and linguistic components on the output probability distribution by an automated comprehension system.    
 
Sentiment classification (SC) is a common NLP classification task where the dominant sentiment class must be selected from a discrete set of sentiments based on a block of input text. This is an example of a single input text element NLP task. The information-theoretic approach is applied here to understand the role of the semantic content and the linguistic realization on the output distribution over the sentiment classes for common sentiment classification datasets. It is interesting to analyze SC as ideally the sentiment of a text block should be based on only its semantic meaning. Here, we determine whether this ideal is held in practice for popular SC corpuses.  


This work makes the following contributions:
\begin{itemize}
    \item Propose an information-theoretic framework for determining the contribution of each text element and further each elemental component on the output distribution for NLP classification tasks.
    \item Detailed analysis of the element and component breakdown according to the proposed framework for multiple-choice reading comprehension and sentiment classification datasets.
\end{itemize}
Despite the framework being applied to NLP classification tasks, it can be adapted to regression, sequence output and even vision tasks.

\section{Related Work}

Features or variables are separate properties that are commonly input to tabular machine learning models to predict a target variable \citep{hwang2023recent}. Feature importance is an active area of research \citep{huang2023feature} where the influence of each feature on the output variable is determined. The ability to determine which features are most important is useful across many verticals e.g. computer assisted medical diagnosis \citep{rudin2019stop}, weather forecasting \citep{malinin2021shifts}, fraud detection \citep{xu2023efficient} and customer churn prediction \citep{alshourbaji2023efficient}. Similarly, we explore the importance of different aspects (can be interpreted as features) at the input including individual elements and the semantic vs linguistic components for NLP text classification tasks. Typically, the structured nature of tabular data allows common feature selection algorithms to be applied including LASSO \citep{tibshirani1996regression}, marginal screening \citep{fan2008sure}, orthogonal matching pursuit \citep{pati1993orthogonal} and decision tree based \citep{costa2023recent}. Due to the relatively unstructured nature of text data (compared to tabular data), we propose an information-theoretic approach to identify the most influential input components.

In multiple-choice reading comprehension, the influence of each element on the final output distribution is directly linked to the complexity of a multiple-choice question. More complex multiple-choice questions can expect to have flat distributions over the answer options while easier questions are sharp around the correct answer. Numerous studies have looked at the factors that potentially influence complexity of context passages in relation with reading comprehension tasks. \citet{sugawara2022makes} observed that the diversity of contexts in MC questions determines the diversity of questions possible conditioned on the contexts.
In their work they found that variables such as passage source, length, or readability measures do not significantly affect the model performance.
Further, \citet{khashabi2018looking} introduced the role of the original source from which the contexts are extracted in shaping overall complexity.
Through the augmentation of the dataset by diversifying the corpus sources, they aimed to enhance the dataset quality.

Question complexity has repeatedly been discussed within prior literature, with varying definitions. \citet{liang2019new} classified questions into distinct categories with complexity scores ranking from lowest to highest for word matching, paraphrasing, single-sentence reasoning, multi-sentence reasoning, and ambiguous questions. \citet{gao2018difficulty} quantified complexity as the number of reasoning steps required to derive the answer. Similar definitions have been upheld by \citet{yang2018hotpotqa} and \citet{dua2019drop}, who expanded the understanding of question complexity to encompass not only contextual comprehension but also factors such as the confidence of a pretrained question-answering model.

Distractor (incorrect options for multiple-choice questions) complexity has been less explored. \citet{gao2019generating} determined distractor complexity based on the similarity between distractors and the ground-truth. Employing an n-gram overlap metric, \citet{banerjee2005meteor} introduced a method to assess distractor complexity. \citet{dugan2022feasibility} further dissected distractor complexity, analyzing qualities such as relevance, interpretability and acceptability compared to human markers. 

As not explored in previous MCRC complexity literature, in this work the information-theoretic approach is applied to characterize the influence of each element in a given multiple-choice reading comprehension dataset. Greater the influence of an element, greater the scope to control the complexity of the multiple-choice reading comprehension task.  

In sentiment classification, it is expected the sentiment class should be dependent on the semantic meaning of the text rather than its linguistic realization. However, \citet{liusie2022analyzing} find that \textit{shortcut} systems that have access only to the stopwords in the original text are also able to identify the sentiment class. Hence, they find the stop words chosen in the text do influence the sentiment class, which we express as the specific linguistic realization. \citet{chew2023understanding} further aim to correct for the bias from spurious correlations. In this work, we explicitly quantify the influence of the semantic and linguistic components of the text.

\section{Theory}
\label{sec:theory}
Here, we describe the generalized framework to analyze the influence of different elements in NLP text classification tasks:
\begin{enumerate}
    \item The individual influence of each element in the input on the output class distribution.
    \item The contribution of the semantic content component vs its linguistic realization component for a given element.
\end{enumerate}

Let an NLP task consist of a set of elements, $\{x_1, \hdots, x_N\} = \mathbf{x}$ influencing the output, $y$, such that:
\begin{equation}
    P(y) = \mathbb{E}_{P\left( \mathbf{x} \right)} P\left(y | \mathbf{x} \right)
\end{equation}

Let $\mathbf{X}$ denote the random variables of each the corresponding instances $\mathbf{x}$. Similarly, let $Y$ be the random variable for an instance of the output, $y$.

To measure the influence of input $\mathbf{x}$ on output $y$, a good metric is the mutual information \citep{depeweg2018decomposition, malinin2018predictive} which measures how the output changes due to variation in the input. Thus we can define $\mathcal{I}(Y;\mathbf{X})$ a measure of the total input influence. 

Similarly we can define the influence from an individual element, $X_j$, $\mathcal{I}(Y;X_j)$ and it should obey:
\begin{equation}
\label{eq:total}
\begin{split}
     \underbrace{\mathcal{I}(Y;X_j)}_{\text{element}} = \underbrace{\mathcal{I}\left(Y;\mathbf{X}\right)}_{\text{total}} -  \underbrace{\mathcal{I}\left(Y; 
\mathbf{X} \textbackslash X_j | X_j \right)}_{\text{other}} 
\end{split}    
\end{equation}
For each element $X_j$, its influence is always determined by two components: $X_j^{(s)}$, the semantic information and a relating linguistic realization method which turns an abstract meaning into natural language.
Thus, we can calculate the semantic influence as $\mathcal{I}(Y;X_j^{(s)})$ and the linguistic influence implicitly $\mathcal{I}(Y;X_j|X_j^{(s)})$. They should satisfy the following relation:
\begin{equation}
\label{eq:component}
\begin{split}
     \underbrace{\mathcal{I}(Y;X_j)}_{\text{element}} = \underbrace{\mathcal{I}\left(Y;X_j^{(s)}\right)}_{\text{semantic}} +\underbrace{\mathcal{I}\left(Y;X_j|X_j^{(s)}\right)}_{\text{linguistic}}
\end{split}    
\end{equation}
In practice for an element, $x_j$, its semantic content is too abstract to be available. Instead we get access to one of its realization ${\tilde{r}}_j$ which is considered to be generated from its unobserved semantic content, $x^{(s)}_j$.
A set of possible realizations of this semantic element, ${\cal R}^{(i)}$, are additionally 
where each member of this
set is, ${r}^{(i)}_j$ drawn as
\begin{eqnarray}
\label{eq:sample}
{r}^{(i)}_j \sim P_r(r|{\tilde r}_i) \approx P_r(r|s_i)
\end{eqnarray}
With these settings, the influence of each component is calculated as follows.

The total influence is:
\begin{eqnarray}
\label{eq:total_entropy}
    \lefteqn{
    \mathcal{I}(Y;\mathbf{X}) } \\
    &=&{\cal H}\left(\mathbb{E}_{P(\mathbf{x})}[P(y|\mathbf{x})]\right)- \mathbb{E}_{P(\mathbf{x})}[{\cal H}(P(y|\mathbf{x}))]\nonumber
     \end{eqnarray}

     
We can also get the element influence as :
\begin{eqnarray}
    \lefteqn{\mathcal{I}(Y;X_j) } \\
    &=&{\cal H}\left(\mathbb{E}_{P(\mathbf{x})}[P(y|\mathbf{x})]\right)- \mathbb{E}_{P(x_j)}\left[{\cal H}(P(y|x_j))\right]\nonumber
    \end{eqnarray}
Further it can be decomposed as the semantic influence:
\begin{eqnarray}
    \mathcal{I}\left(Y;X_j^{(s)}\right)
    \!\!\!&=&\!\!\!{\cal H}\left(\mathbb{E}_{P(\mathbf{x})}[P(y|\mathbf{x})]\right)\\
    \!\!\!&
    -&\!\!\!\mathbb{E}_{P\left( x_j^{(s)} \right)}\left[{\cal H}\left(P\left(y| x_j^{(s)} \right)\right)\right]
    \nonumber
    \end{eqnarray}
and the linguistic influence:
\begin{eqnarray}
    \mathcal{I}\left(Y;X_j|X_j^{(s)}\right)
    \!\!\!&=&\!\!\!\mathbb{E}_{P\left(x_j^{(s)}\right)}
    \left[{\cal H}\left(P\left(y|x_j^{(s)}\right)\right)\right] \nonumber \\
    \!\!\!&  -& \!\!\!
    \mathbb{E}_{P(x_j)}\left[{\cal H}\left(P(y|x_j)\right)\right]
    \label{eq:linguistic}
    \end{eqnarray}
The relative contribution of an element to the total influence and the relative contribution of the semantic component for an element can respectively be expressed as:
\begin{equation}
    \text{relative element influence } = \frac{\mathcal{I}(Y;X_j)}{\mathcal{I}(Y;\mathbf{X})}
\end{equation}
\begin{equation}
    \text{relative semantic influence } = \frac{\mathcal{I}\left(Y;X_j^{(s)}\right)}{\mathcal{I}(Y;X_j)}
\end{equation}
    
\subsection{Multiple-choice reading comprehension}
\label{sec:theory-mcrc}
The technique of multiple-choice reading comprehension is widely used to evaluate candidates' reading comprehension skills in standardized tests \citep{frizelle_o'neill_bishop_2017}. In this task, candidates are provided with a context passage, $c$ and a corresponding question, $q$. The objective is to determine the correct answer from a defined set of options, denoted as $o$. This process involves understanding the question and utilizing the context passage as a source of information to ascertain the most appropriate answer option.

The output distribution for a reading comprehension question can be categorised as:
\begin{equation}
\label{eq:mrc_abstract}
    P(y) = \mathbb{E}_{P(c,q,o)} P(y | c,q,o)    
\end{equation}

    
\subsubsection{Data generation}
\begin{figure}
    \centering
    \includegraphics[width=1.0\linewidth]{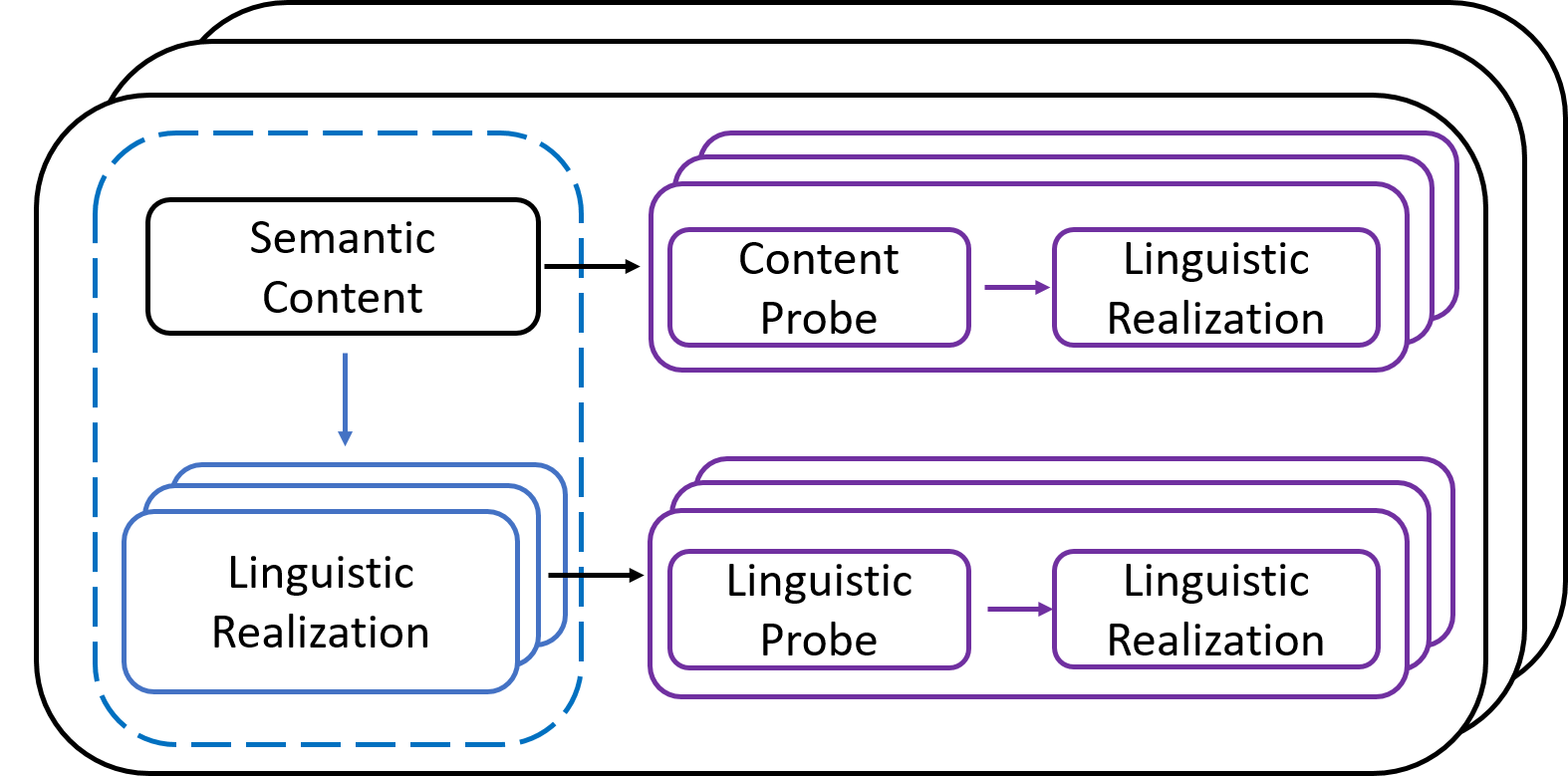}
    \caption{Data generation for multiple-choice reading comprehension for the context (blue) and question (purple) respectively.}
\label{fig:mcrc_illustration}
\end{figure}

For a typical MRC dataset, the data generation process is shown in Figure \ref{fig:mcrc_illustration}. The generation of a context is shown in the dotted blue block: a specific semantic content $c^{(s)}$ is chosen and a context $c$ is generated when a certain linguistic realization $r$ is applied. Therefore, the influence of the context $C$ can be divided into $\mathcal{I}(Y;C^{(s)}), \mathcal{I}(Y;C|C^{(s)})$. The influence of the context realization appears implicitly as it controls $\mathcal{I}(Y;C|C^{(s)})$

A similar procedure is applied on the questions and the options but usually (for the scope of this work) they are generated together as a question-option pair $q$: from a certain semantic content, a content probe is generated and then a question-option pair in natural language is a linguistic realization of this probe as described by: $P(q|c^{(s)})$.

Thus, the output distribution for a reading comprehension question can be rewritten as:
\begin{equation}
\begin{split}
     P(y) = \mathbb{E}_{P(c^{(s)})}\mathbb{E}_{P(c|c^{(s)})}\mathbb{E}_{P(q|c^{(s)})}P(y|q, c)
\end{split}    
\end{equation}

If we consider only the questions generated from the semantic contents and ignore the questions constrained to a specific realization by filtering out all questions generated from the linguistic content of the context, we can rewrite the generation process of a question as:
$P(q|c)$.
To make this assumption valid, we need to filter out all linguistic questions. The specific method is shown in Appendix \ref{app:question_filter}. Thus Equation \ref{eq:mrc_abstract} is rewritten as:
\begin{equation}
\label{eq:mcrc_break}
\begin{split}
     P(y) = \mathbb{E}_{P(c^{(s)})}\mathbb{E}_{P(c|c^{(s)})}\mathbb{E}_{P(q|c)}P(y|q, c)
\end{split}    
\end{equation}
Note, different context realizations are generated as paraphrases conditional on the original context such that $r \sim P_{\text{gpt}}(r|c)$.


\subsubsection{Measure of component influence}
\label{sec:theory-comp-inf}
The question-option pair in Equation \ref{eq:mcrc_break} appear as $P(q|c)$, thus instead of $\mathcal{I}(Y;Q)$, we consider $\mathcal{I}(Y;Q|C)$ and get the decomposition of total influence following Equation \ref{eq:total}:
\begin{equation}
\label{eq:mc_total}
\begin{split}
 \underbrace{\mathcal{I}(Y;C)}_{\text{context}}  = \underbrace{\mathcal{I}(Y;C,Q)}_{\text{total}} -  \underbrace{\mathcal{I}(Y;Q|C)}_{\text{question}} 
\end{split}    
\end{equation}
The context influence can be further decomposed according to Equation \ref{eq:component}:
\begin{equation}
\label{eq:mc_component}
\begin{split}
     \underbrace{\mathcal{I}(Y;C)}_{\text{context}} = \underbrace{\mathcal{I}(Y;C^{(s)})}_{\text{semantic}} +\underbrace{\mathcal{I}(Y;C|C^{(s)})}_{\text{linguistic}}
\end{split}    
\end{equation}

It is clear that Equations \ref{eq:mc_total} and \ref{eq:mc_component} are an example of Equations \ref{eq:total} and \ref{eq:component} respectively. Thus similar to Section \ref{sec:theory}, the influence terms can be calculated according to Equations \ref{eq:total_entropy} to \ref{eq:linguistic}.
Besides the assumption made in Equation \ref{eq:sample} which is general for all the tasks, a further assumption about the questions are made for the multiple-choice reading comprehension task: instead of sampling from the ideal question generation process, we only observe the question-option pairs generated by humans, ${\tilde{\cal Q}}^{(i)}$, where each member of this set is, ${\tilde q}^{(i)}_j$ drawn as:
\begin{eqnarray}
{\tilde q}^{(i)}_j \sim P_{\text{man}}(q|{\tilde r}_i)  \approx P_q\left(q|c_i^{(s)}\right)
\end{eqnarray}
It is then possible to approximate the total influence using:
\begin{eqnarray}
\lefteqn{\mathbb{E}_{P(c,q)}\left[P(y|c,q)\right] \approx}\\
&&\frac{1}{n_{\tt s}}\sum_{i=1}^{n_{\tt s}}\frac{1}{|{\cal R}^{(i)}||{\tilde{\cal Q}}^{(i)}|}\sum_{r\in {\cal R}^{(i)},{\tilde q}\in{\tilde{\cal Q}}^{(i)}} 
\!\!\!\!P(y|{\tilde q},r)\nonumber
\end{eqnarray}
and
\begin{eqnarray}
\lefteqn{\mathbb{E}_{P(c,q)}\left[{\cal H}(P(y|c,q))\right]\approx}\\
&&\frac{1}{n_{\tt s}}\sum_{i=1}^{n_{\tt s}}\frac{1}{|{\cal R}^{(i)}||{\tilde{\cal Q}}^{(i)}|}
\sum_{r\in {\cal R}^{(i)},{\tilde q}\in{\tilde{\cal Q}}^{(i)}} 
\!\!\!\!{\cal H}(P(y|{\tilde q},r))\nonumber
\end{eqnarray}
and
\begin{eqnarray}
\lefteqn{\mathbb{E}_{P(c)}\left[{\cal H}(P(y|c))\right]\approx}\\
&&\frac{1}{n_{\tt s}}\sum_{i=1}^{n_{\tt s}}\frac{1}{|{\cal R}^{(i)}|}
\sum_{r\in {\cal R}^{(i)}} 
\!{\cal H}(\!\frac{1}{|{\tilde{\cal Q}}^{(i)}|}\sum_{{\tilde q}\in{\tilde{\cal Q}}^{(i)}}P(y|{\tilde q},r))\nonumber
\end{eqnarray}
and
\begin{eqnarray}
\lefteqn{\mathbb{E}_{P(c^{(s)})}\left[{\cal H}(P(y|c^{(s)}))\right]\approx}\\
&&\frac{1}{n_{\tt s}}\sum_{i=1}^{n_{\tt s}} 
\!{\cal H}(\frac{1}{|{\cal R}^{(i)}|}
\sum_{r\in {\cal R}^{(i)}}\!\frac{1}{|{\tilde{\cal Q}}^{(i)}|}\sum_{{\tilde q}\in{\tilde{\cal Q}}^{(i)}}P(y|{\tilde q},r))\nonumber
\end{eqnarray}
with $n_{s}$ is the number of contexts in a dataset.


\subsection{Sentiment classification}

For the sentiment classification task, the candidate receives a sentence or a short paragraph $x$ and then is requested to choose the sentiment class that describes the input best from the set of sentiment options $o$. 
\begin{equation}
    P(y) = \mathbb{E}_{P(x,o)}P(y|x,o)
\end{equation}

In sentiment classification, the sentiment classes (unlike the options in multiple-choice reading comprehension where the associated text with each option class changes across different examples) are fixed. Hence, we can drop the dependence on the sentiment options. 

Here we are only interested the influence to the output $y$ from semantic content $x^{(s)}$ and its linguistic realization method: $\mathcal{I}(Y;X^{(s)}), \mathcal{I}(Y;X|X^{(s)})$, as there is only one element at the input. Following Equation \ref{eq:component}, the semantic and linguistic breakdown is expressed as:
\begin{equation}
\label{eq:sc_component}
\begin{split}
     \underbrace{\mathcal{I}(Y;X)}_{\text{text}} = \underbrace{\mathcal{I}(Y;X^{(s)})}_{\text{semantic}} +\underbrace{\mathcal{I}(Y;X|X^{(s)})}_{\text{linguistic}}
\end{split}    
\end{equation}
In practice, the following approximations are made:

\begin{eqnarray}
\lefteqn{\mathbb{E}_{P(x)}\left[{\cal H}(P(y|x))\right]\approx}\\
&&\frac{1}{n_{\tt s}}\sum_{i=1}^{n_{\tt s}}\frac{1}{|{\cal R}^{(i)}|}
\sum_{r\in {\cal R}^{(i)}} 
\!{\cal H}(P(y|r))\nonumber
\end{eqnarray}
and
\begin{eqnarray}
\lefteqn{\mathbb{E}_{P(x^{(s)})}\left[{\cal H}(P(y|x^{(s)}))\right]\approx}\\
&&\frac{1}{n_{\tt s}}\sum_{i=1}^{n_{\tt s}} 
\!{\cal H}(\frac{1}{|{\cal R}^{(i)}|}
\sum_{r\in {\cal R}^{(i)}}P(y|r))\nonumber
\end{eqnarray}

\section{Systems}

\begin{figure*}[t]
    \centering
        \includegraphics[width=1.0\linewidth]{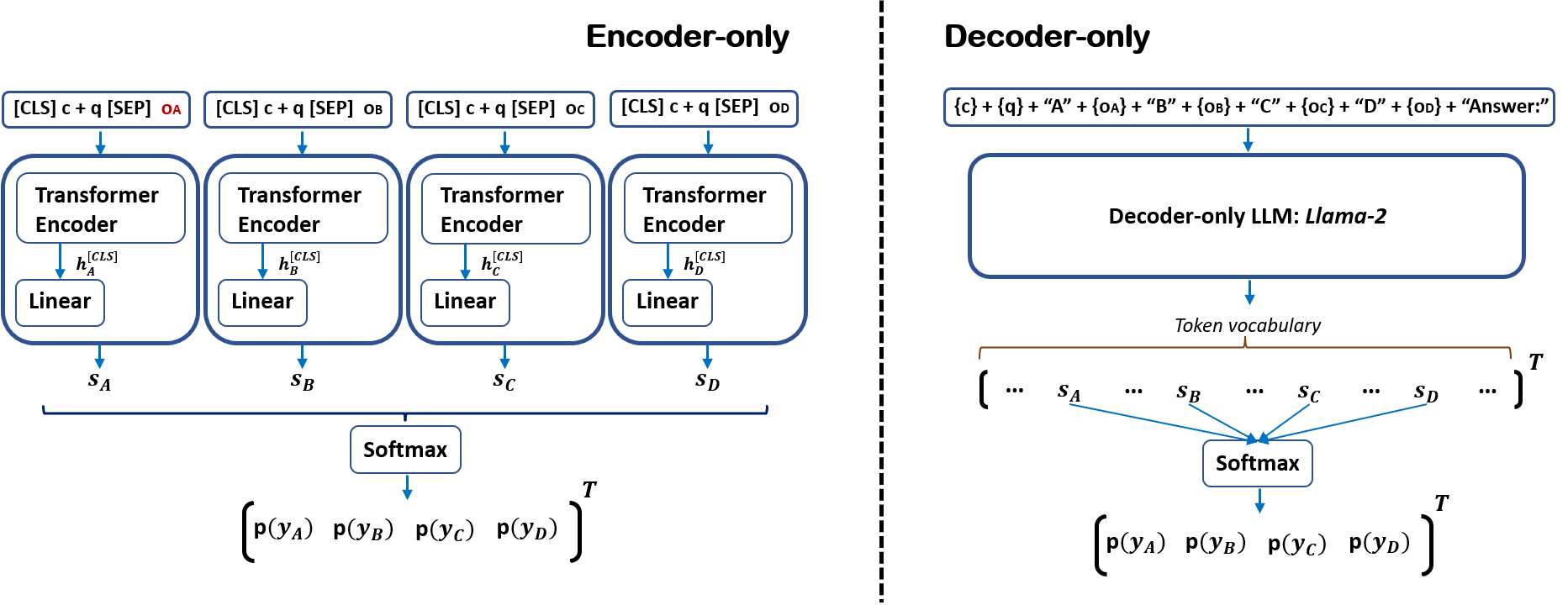}
    \caption{Architectures for multiple-choice reading comprehension with context, $c$, question, $q$ and options, $o$.}
    \label{fig:arch_comprehension}
\end{figure*}

\subsection{Linguistic realization}
\label{sec:linguistic}

In order to analyze the impact of the linguistic realization of a given text element, it is necessary to fix the semantic content of the element. 

In other work (such as \citet{sugawara2022makes}), there has been an attempt evaluate the effect of linguistic content of the context for multiple-choice reading comprehension. However, they ignore the requirement to fix the semantic content to fairly measure the impact of different linguistic realizations.

In this work, we employ a paraphrasing system to generate different linguistic realizations for the same semantic content of a text element. The paraphrasing approach is applied to the context element in multiple-choice reading comprehension and to the input element in sentiment classification.

In order to consider a broad range of linguistic realizations for a specific text's semantic content, we generate 8 paraphrases at different readability levels. Hence, we assume (this assumption is assessed in Appendix \ref{app:paraphrasing}) that the linguistic realizations at different readability levels maintain the same semantic content.

To change the readability of the text element, we use the zero-shot method as in \citet{farajidizaji2023possible} based on Equation \ref{eq:sample}:
\begin{equation}
    {r}^{(i)}_j \sim P_{\text{LLM}}(r|{\tilde r}_i)
\end{equation}

In practice, the zero-shot large language model (LLM), in this work ChatGPT 3.5 \footnote{Available at: \url{https://openai.com/pricing}}, is fed with the original text along with an instruction to alter the language of the text to match the desired readability level. The model, not previously trained on this specific task, uses its extensive pre-existing knowledge and understanding of language structure and complexity to alter the readability whilst maintaining the same semantic meaning. 

The readability level is measured by Flesch reading-ease \citep{flesch1948new} score (FRES) as shown below, where higher scores
indicate material that is easier to read while lower scores are reflective of more challenging texts.
$$
   {\tt FRES} = \text{206.835} - \text{1.015}\left(\frac{n_{w}}{n_{se}}\right) - \text{84.6}\left(\frac{n_{sy}}{n_{w}}\right) 
$$
$n_{w}$ is the total number of words, $n_{se}$ is the total number of sentences, $n_{sy}$ the total number of syllables.

The prompts to generate paraphrases at different readability levels for a given input element are indicated in Table \ref{tab:para_prompts} in Appendix \ref{app:paraphrasing}.

\subsection{Reading comprehension}

In this work, multiple-choice reading comprehension systems are required to return a probability distribution over the answer options. Two alternative architectures are considered for performing the reading comprehension task.

Figure \ref{fig:arch_comprehension} presents the standard encoder-only architecture \citep{Yu2020ReClorAR, raina-gales-2022-answer, liusie2023world, raina2023analyzing} for multiple-choice reading comprehension systems used in this work for questions involving 4 options. Each option is individually encoded along with both the question and the context to produce a score. A softmax function is then applied to the scores linked to each option, transforming them into a probability distribution. During the inference phase, the anticipated answer is chosen as the option with the highest associated probability. The parameters of the transformer encoder \citep{vaswani2017attention} and the linear layer are shared for all options, allowing the number of options to be flexible.

Inspired by \citet{liusie2023zero} and the recent success observed in finetuning large open-source instruction finetuned language models \citep{touvron2023llama, touvron2023llama2, jiang2023mistral,jiang2024mixtral, tunstall2023zephyr} on various NLP tasks, this work additionally finetunes Llama-2 \citep{touvron2023llama2} for multiple-choice reading comprehension. The approach is also presented in Figure \ref{fig:arch_comprehension}. The context, question and answer options are concatenated into a single natural language prompt.
As an autoregressive language model, Llama-2 is requested to effectively return a single token at the output, represented by a single logit distribution over the token vocabulary. The logits associated with the tokens \textit{A,B,C,D} are respectively noramlized using softmax to return the desired probability distribution over the answer options. As with the encoder-only architecture, the option with the highest probability is selected as the answer at inference time.

\subsubsection{Calibration}

The trained models were calibrated post-hoc using single parameter temperature annealing \citep{guo2017calibration}. It is necessary to calibrate the models for the absolute information-theoretic measures to be meaningful.

Uncalibrated, model probabilities are determined by applying the softmax to the output logit scores $s_i$:
\begin{equation}
    P(y = k ; \bm{\theta}) \propto \exp(s_k)
\end{equation}
where $k$ denotes a possible output class for a prediction $y$.
Temperature annealing `softens' the output probability distribution by dividing all logits by a single parameter $T$ prior to the softmax. 
\begin{equation}
    P_{CAL}(y = k ; \bm{\theta}) \propto \exp(s_k/T)
\end{equation}
As the parameter $T$ does not alter the relative rankings of the logits, the model's prediction will be unchanged and so temperature scaling does not affect the model’s accuracy. The parameter $T$ is chosen such that the accuracy of the system is equal to the mean of the maximum probability (as is expected for a calibrated system).


\subsection{Data complexity classification}
\label{sec:data_comp_model}
\begin{figure}[htbp!]
    \centering
    \includegraphics[width=0.8\linewidth]{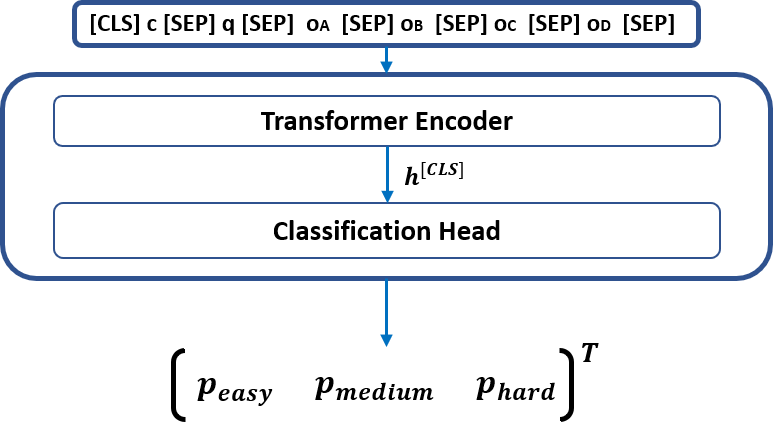}
    \caption{Architecture for MC question complexity classifier with context, $c$, question, $q$ and options, $\{o\}$.}
    \label{fig:arch_comp}
\end{figure}

Here, an automated complexity system takes all the components of an MC question and classifies it into one of the 3 classes: \textit{easy}, \textit{medium} or \textit{hard}.
Figure \ref{fig:arch_comp} presents the standard architecture used for MC question complexity classification \citep{raina2022multiple, benedetto2023quantitative}. The input consists of all components of an MC question concatenated together which are fed into a transformer encoder architecture. The hidden embedding representation of the prepended \texttt{[CLS]} token is taken as the sentence embedding representation of the MC question, which in turn is passed to a classification head. The head, which is a single linear layer, returns a probability distribution over the three complexity levels.

In order to empirically investigate the relative importance of each component in an MC question, various input formats are trialled in the question complexity system. Table \ref{tab:inputs} presents different combinations of the context, question and answer options where one or more components have been omitted from the standard (full) input format. Each input is prepended with the \texttt{[CLS]} token and each component is separated by the \texttt{[SEP]} token.

\begin{table}[htbp!]
\centering
\small 
\begin{tabular}{ll}
\toprule
standard & \texttt{c + q + $\texttt{o}_\texttt{A}$ + $\texttt{o}_\texttt{B}$ + $\texttt{o}_\texttt{C}$ + $\texttt{o}_\texttt{D}$} \\
context & \texttt{c} \\
context-question & \texttt{c + q} \\
question-option & \texttt{q + $\texttt{o}_\texttt{A}$ + $\texttt{o}_\texttt{B}$ + $\texttt{o}_\texttt{C}$ + $\texttt{o}_\texttt{D}$} \\
   \bottomrule
    \end{tabular}
\caption{Input formats for the complexity system.}
\label{tab:inputs}
\end{table}

\subsection{Sentiment classification}

Sentiment classification models take the input text and return a probability distribution over the set of sentiments. In this work, the sentiments for the datasets considered are \{negative, positive \}.

As a standard NLP text classification task, we take the standard approach of taking a pretrained transformer encoder model \citep{vaswani2017attention} with a classification head at the output \citep{liusie2022analyzing}.
Similar to the encoder-only approach for multiple-choice reading comprehension (Figure \ref{fig:arch_comprehension}) and the data complexity classification system (Figure \ref{fig:arch_comp}), the sentiment classification system only passes the hidden embedding representation of the \texttt{[CLS]} token to the classification head. A softmax function is applied to normalize the logits into a probability distribution over the sentiment classes.

\section{Experiments}

\subsection{Data}

\begin{table*}[htbp!]
\small
    \centering
    \begin{tabular}{ll|cccccc}
        \toprule
& & \# examples & \# options & \# words & \# questions & semantic diversity & linguistic diversity \\
\midrule
\multirow{3}{*}{RC} 
& MCTest & 142& 4 & 209& 4 & $\text{0.079}_{\pm{\text{0.015}}}$ & $\text{0.018}_{\pm{\text{0.006}}}$\\
& RACE++ & 1,007 & 4 & 278 & 3.7& $\text{0.101}_{\pm{\text{0.015}}}$ & $\text{0.016}_{\pm{\text{0.007}}}$\\
& CMCQRD & 150& 4 & 683& 5.5 &  $\text{0.092}_{\pm{\text{0.010}}}$& $\text{0.022}_{\pm{\text{0.011}}}$\\
\midrule
\multirow{3}{*}{SC} & IMDB & 500 & 2 & 226 & - & $\text{0.084}_{\pm{\text{0.019}}}$ & $\text{0.023}_{\pm{\text{0.011}}}$\\
& Yelp & 500 & 2 & 133 & - & $\text{0.108}_{\pm\text{0.037}}$ & $\text{0.030}_{\pm\text{0.024}}$\\
& Amazon & 500 & 2 & 74 & - & $\text{0.135}_{\pm\text{0.024}}$& $\text{0.037}_{\pm\text{0.022}}$\\
        \bottomrule
    \end{tabular}
    \caption{Statistics for reading comprehension (RC) and sentiment classification (SC) test datasets. See Appendix \ref{tab:app_data_stats} for additional datasets.}
    \label{tab:test_data}
\end{table*}

We use the RACE++ reading comprehension dataset \citep{lai2017race, pmlr-v101-liang19a} train split for training both the MC data complexity evaluator and the MC reading comprehension model. The RACE++ dataset is the largest publicly available dataset from English exams in China partitioned into three difficulty levels: middle school (RACE-M), high school (RACE-H) and college (RACE-C) in increasing order of difficulty. The detailed information about the splits of RACE++ is given in Appendix \ref{app:data}.




Additionally, various multiple-choice reading comprehension datasets are considered as test sets for investigating the influence of each component. Specifically, the test sets are taken from RACE++, MCTest \citep{richardson2013mctest} and CMCQRD \citep{CMCQRD-2023}. The statistics of each dataset is presented in Table \ref{tab:test_data}. MCTest requires machines to answer multiple-choice reading comprehension questions about fictional stories, which addresses the goal of open-domain machine comprehension. CMCQRD is a small-scale multiple-choice reading comprehension dataset from the pre-testing stage partitioned into grade levels B1 to C2 on the Common European Framework of Reference for Languages (CEFR) scale.

For sentiment classification, this work considers the IMDb \citep{maas2011learning},
Yelp-polarity (Yelp) \citep{zhang2015character} and Amazon-polarity (Amazon) \citep{mcauley2013hidden}
datasets.
IMDb is a binary sentiment analysis dataset consisting of reviews from the Internet Movie Database.
Yelp consists of reviews from Yelp where 1 or 2 stars is interpreted as negative while 4 or 5 stars is interpreted as positive.
Amazon consists of reviews from Amazon over a period of 13 years on various products. Both Amazon and Yelp are hence binary sentiment classification datasets.

Table \ref{tab:test_data} details the main statistics for each of these test sets. All the reading comprehension test sets have 4 options for classification. The selected sentiment classification test sets have 2 options: negative and positive. The number of words for reading comprehension refers to the lengths of the contexts. It is seen that the test sets considered have varying lengths from 200 to 700 words and 75 to 230 words for reading comprehension and sentiment classification tasks respectively.

For the reading comprehension datasets, the total number of examples are reported after filtering out all linguistic probe questions (see Appendix \ref{app:question_filter} for the procedure). So the focus is only on the semantic probe questions as assumed in Section \ref{sec:theory-comp-inf}. For the sentiment classification test sets, a subset of 500 examples is selected for each dataset to remain within the financial budget for use of the ChatGPT API for the generation of different linguistic realizations as paraphrases.

Additionally, the semantic diversity is calculated for each dataset. As explored in \citet{raina-etal-2023-erate}, the semantic diversity score is the mean cosine distance between each text embedding in the dataset to the centroid of all text embedding representations (an approximation of the radial distance in the embedding hyperspace). Greater the semantic diversity score, greater the variation in the set of texts being considered. The sentence embedder based on \citet{ni2022sentence} is used to generate embeddings for each text \footnote{Available at: \url{https://huggingface.co/sentence-transformers/sentence-t5-base}}. Note, the semantic diversity is calculated on the contexts for the reading comprehension datasets. Finally, the linguistic diversity calculates the mean variation in the embedding space for different linguistic realizations (by paraphrasing - see Section \ref{sec:linguistic}) for each context. 

{
\centering
\begin{table*}[h]
\small
    \centering
    \begin{tabular}{l|cc|ccc|cc}
    \toprule
    \multirow{2}{*}{dataset}& \multicolumn{2}{c|}{accuracy} & \multicolumn{5}{c}{influence}\\
     & original & para & total & question & context  & context-semantic & context-linguistic \\
    \midrule
MCTest & 92.5 & 85.8 & 0.212 & 0.116 (54.7\%) & 0.096 (45.3\%) & 0.068 (70.6\%) & 0.028 (29.4\%) \\
RACE++ & 86.0& 82.9 & 0.298 &0.161 (56.1\%)&0.131 (43.9\%)&0.108 (82.5\%)& 0.023 (17.5\%)\\
CMCQRD & 79.9 & 69.4 & 0.290 & 0.211 (72.7\%) & 0.079 (27.3\%) & 0.067 (83.8\%) & 0.012 (16.2\%) \\
    \bottomrule
    \end{tabular}
    \caption{Decomposition of total input influence for Llama-2 on reading comprehension datasets.}
    \label{tab:test_accs}
\end{table*}
}

\subsection{Model details}
\label{sec:model}

For the standard encoder-only implementations of the multiple-choice reading comprehension system (Figure \ref{fig:arch_comprehension}), two pretrained transformer encoder models are selected as the backbone: RoBERTa \citep{liu2019roberta} and the Longformer \citep{beltagy2020longformer}. Refer to Appendix \ref{app:models} for further details about these models.

The decoder-only implementation of the multiple-choice reading comprehension system is based upon the pretrained instruction finetuned Llama2-7B model \footnote{Available at \url{https://huggingface.co/meta-llama/Llama-2-7b-chat-hf}}. 

ELECTRA-base \citep{clark2020electra} is selected for the data complexity evaluator models. The pretrained model is finetuned on the RACE++ train split with the complexity class (easy, medium or hard) as the label. Hyperparameter tuning details are given in Appendix \ref{app:models} for both the reading comprehension and data complexity classifier systems.


For sentiment classification, the main paper reports results based on a RoBERTa architecture. The selected model has been finetuned on IMDb training data \footnote{Available at \url{https://huggingface.co/wrmurray/roberta-base-finetuned-imdb}}
Due to the similarity in content between Yelp, Amazon and IMDb, the RoBERTa model finetuned on IMDb is also applied on all sentiment classification test sets at inference.
For further reproducibility, a BERT-based \citep{devlin2018bert} system is considered. Model details are given for this system in Appendix \ref{app:models}.

\section{Results}
\label{sec:results}



\subsection{Reading Comprehension}
\label{sec:res-rc}

\begin{figure*}[t]
    \centering
    \begin{subfigure}[t]{1\columnwidth}
        \centering
        \includegraphics[width=3in]{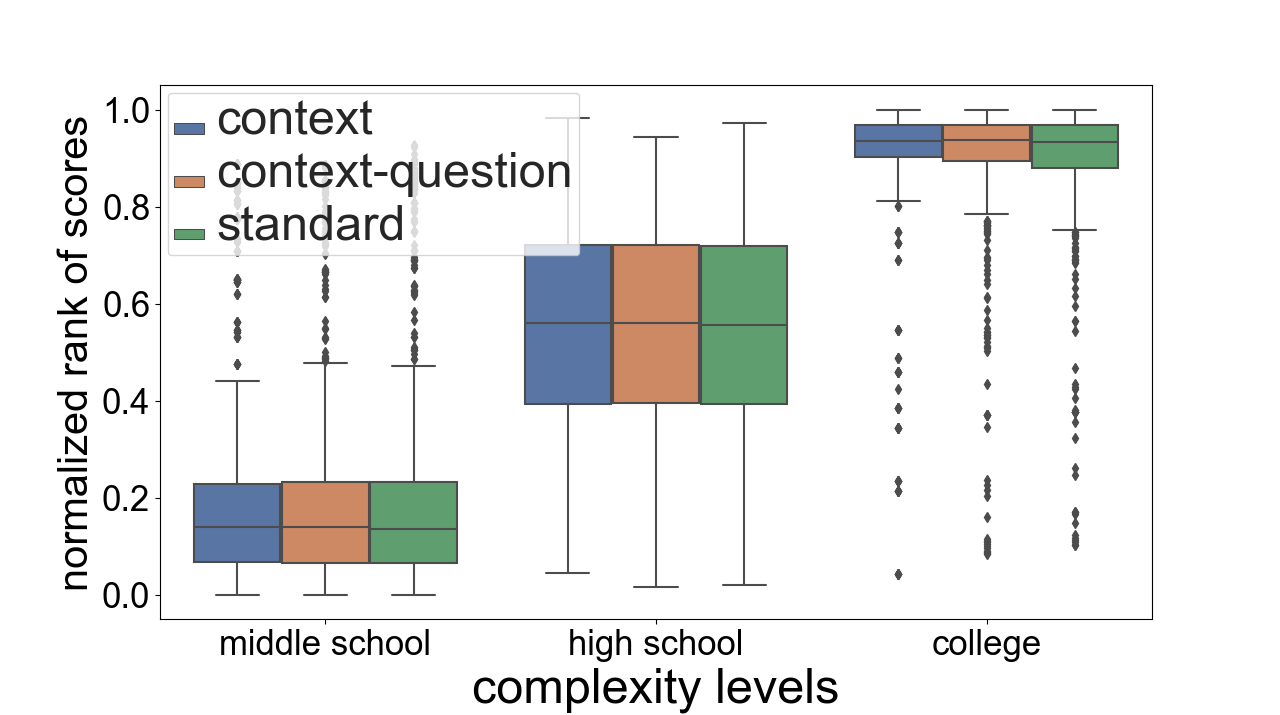}
        \caption{RACE++}
    \end{subfigure}%
    ~ 
    \begin{subfigure}[t]{1\columnwidth}
        \centering
        \includegraphics[width=3in]{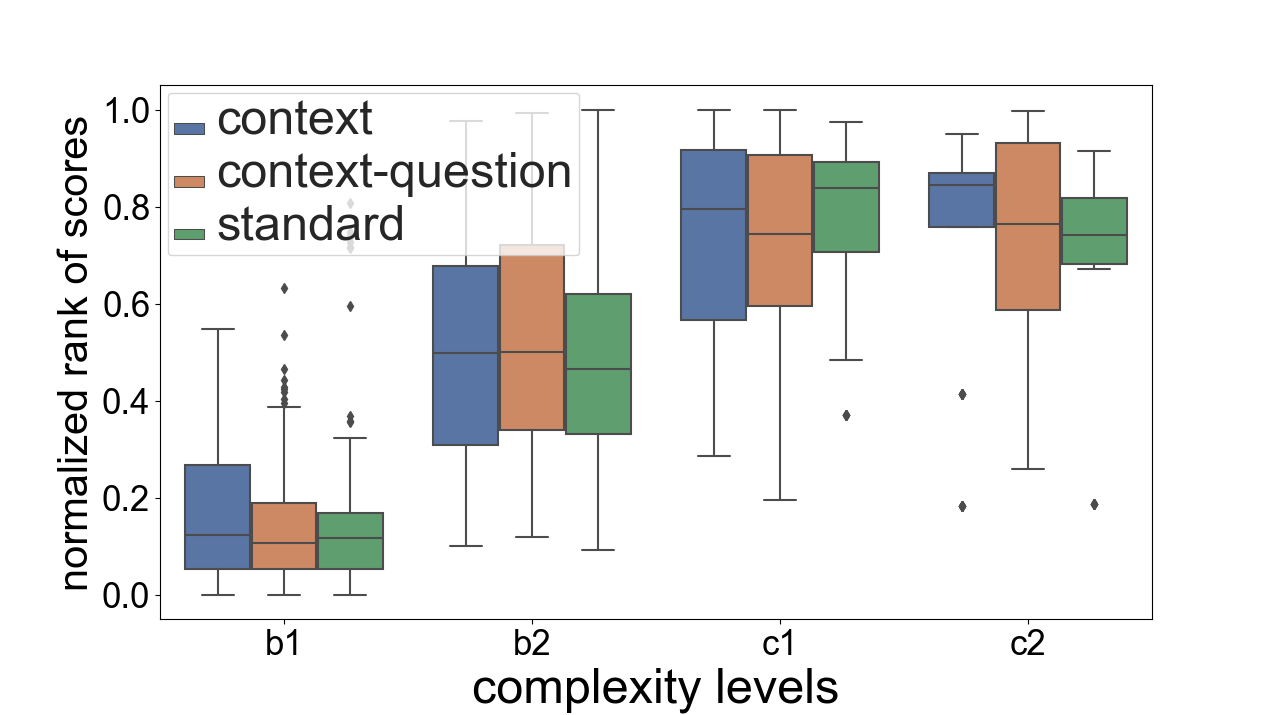}
        \caption{CMCQRD}
    \end{subfigure}%
    \caption{Normalized ranks (rank / total examples) of complexity scores for each complexity level using three complexity evaluators: context, context-question and standard.}
    \label{fig:box_complexities}
\end{figure*}

\begin{figure*}[t]
    \centering
    \begin{subfigure}[t]{1\columnwidth}
        \centering
        \includegraphics[width=3in]{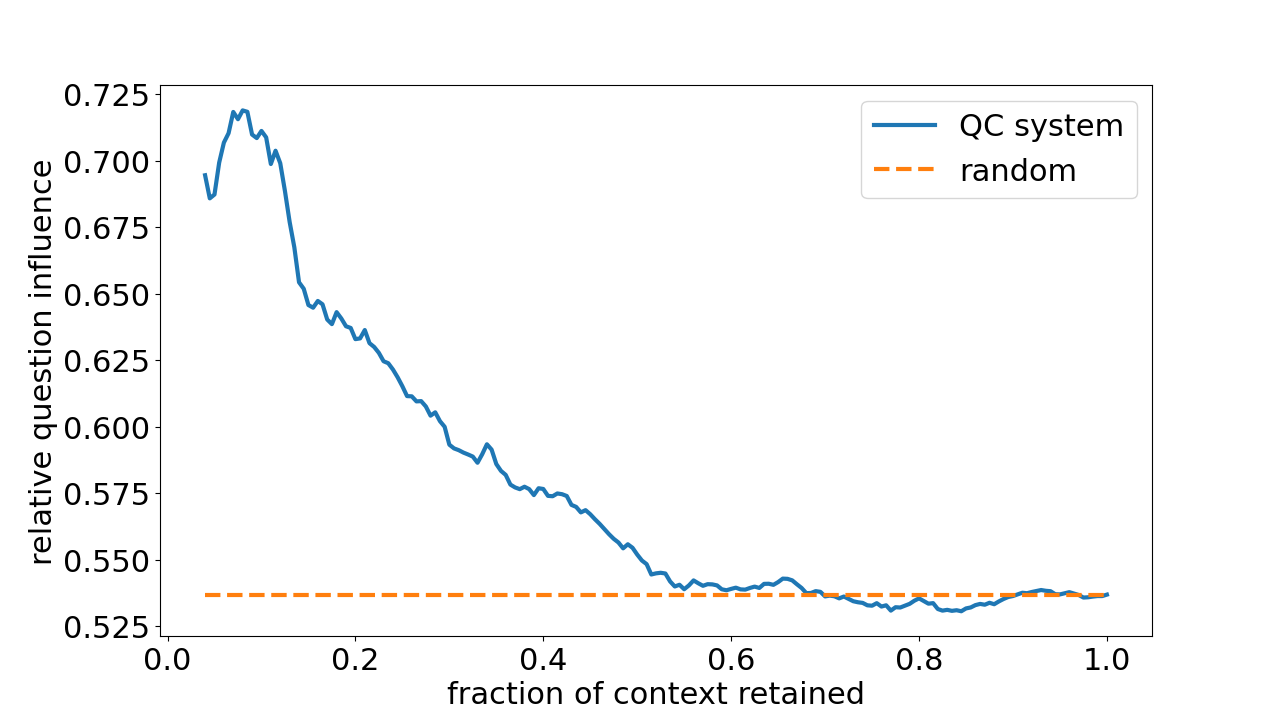}
        \caption{Total}
    \end{subfigure}%
    ~ 
    \begin{subfigure}[t]{1\columnwidth}
        \centering
        \includegraphics[width=3in]{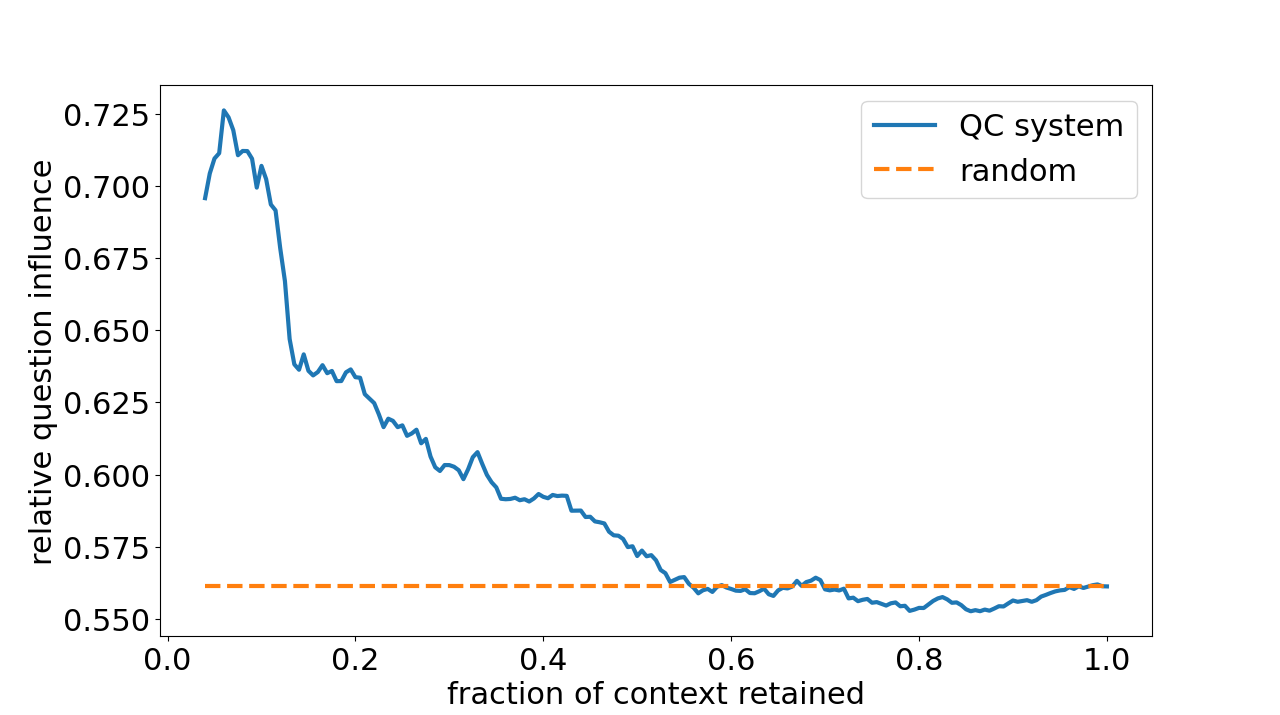}
        \caption{RACE++}
    \end{subfigure}%
    \caption{The relative question influence changes with the subset chosen by the rank of context complexity in all three datasets (left) and in RACE++ only (right). \footnote{The number of examples is to CMCQRD and MCTest dataset is too few. Hence, their figures are not shown here.} 0.2 in x-axis means we leave contexts with top 20\% context complexity as the subset.}
    \label{fig:q_inf_sweep}
\end{figure*}

Table \ref{tab:test_accs} presents the performance of Llama-2 on the various reading comprehension datasets. The highest accuracy is observed on MCTest with 92.5\% and the lowest on CMCQRD with 79.9\%. Additionally, the performance of the model is reported on each dataset after generating 8 paraphrases for each context (see Section \ref{sec:linguistic}). It is observed there is a consistent drop in performance of the model on the paraphrased contexts compared to the original. This is expected as the nature of the machine generated paraphrased contexts do not necessarily align with the type of contexts observed in the original dataset.

Table \ref{tab:test_accs} further investigates the influence of the different elements: specifically the context influence compared to the question influence (note the question includes the question and the options - see Section \ref{sec:theory-mcrc}). It is evident that the context of a multiple-choice reading comprehension question plays an important role in the final output distribution with influences up to 45\% for MCTest. 

The complexity of a multiple-choice reading comprehension question is described by the shape of the output distribution. A sharp distribution about the correct answer is indicative of an easy question while a flatter distribution over all the answer options indicates a harder question. Therefore, the strong influence of the context demonstrated in Table \ref{tab:test_accs} emphasises that the context (alongside the specific posed question) is important in controlling the complexity of a question. 

In order to further verify the influence of the context on the complexity of a multiple-choice reading comprehension question, Figure \ref{fig:box_complexities} plots the complexity score output by the data complexity classifier (see Section \ref{sec:data_comp_model}). In particular, the distribution (as a boxplot) is shown for the complexity scores on the different subsets from RACE++ and CMCQRD of different complexity levels. Note, the noramalized ranks of the complexity scores is plotted where the global rank is found for a given complexity score and divided by the total number of examples. The distributions are shown for the standard system (context, question and options), context-question system (context and question) and the context only system. Figure \ref{fig:box_complexities} clearly shows that the context is sufficient to determine the complexity levels of multiple-choice questions for these datasets, empirically supporting the importance of the context in the complexity of a question.

For the context element, Table \ref{tab:test_accs} further reports the influence for the semantic and linguistic components. For all 3 datasets, the semantic meaning of a context has a greater influence on the final output distribution but the specific linguistic realization of the context also influences the output. Specifically, the relative semantic influence is greatest for MCTest and lowest for CMCQRD.
This is further supported by Table \ref{tab:test_data} where the calculated semantic diversity is the lowest for MCTest with similar linguistic diversities across the datasets. See Appendix \ref{app:data_comp} for the performance of the data complexity classifier system.

Additionally, Table \ref{tab:test_accs} suggests a relationship between the difficulty of a dataset (indicated by the accuracy of the model on the system) and the relative question influence. The relative question influence increases with more challenging datasets. In order to further explore this observation, Figure \ref{fig:q_inf_sweep} determines whether the question influence is directly linked to the complexity of a question. The data complexity classifier (QC system) is used to rank all the contexts according to their complexity.
Then retaining a certain fraction of the most complex contexts, the relative question influence is plotted. The curve is plotted for all the datasets combined and RACE++ on its own. Compared to the random ordering line, it is clear that retaining the most complex contexts leads to larger question influence scores.
The increase in question influence with more challenging contexts supports the trend from Table \ref{tab:test_accs}. Therefore, a more challenging context allows a greater variation in question difficulties, leading to a greater question influence on the output distribution.

{
\centering
\begin{table*}[h]
\small
    \centering
    \begin{tabular}{c|cc|ccc}
    \toprule
    \multirow{2}{*}{dataset}& \multicolumn{2}{c|}{accuracy} & \multicolumn{3}{c}{influence}\\
    & original & para & total & context-semantic & context-linguistic \\
    \midrule
     IMDB &  94.8& 93.4& 0.472 & 0.444 (94.0\%)& 0.028 (6.0\%) \\
    Yelp & 94.3 & 93.9 & 0.472 &  0.445 (94.2\%)&  0.027 (5.8\%)\\
    Amazon &  91.0&  89.5 & 0.361 & 0.325 (90.0\%)& 0.036 (10.0\%)\\
    \bottomrule
    \end{tabular}
    \caption{Decomposition of context influence for different models in various datasets for sentiment classification.}
    \label{tab:test_accs_sent}
\end{table*}
}

\begin{figure*}[t]
    \centering
    \begin{subfigure}[t]{1\columnwidth}
        \centering        \includegraphics[width=3in]{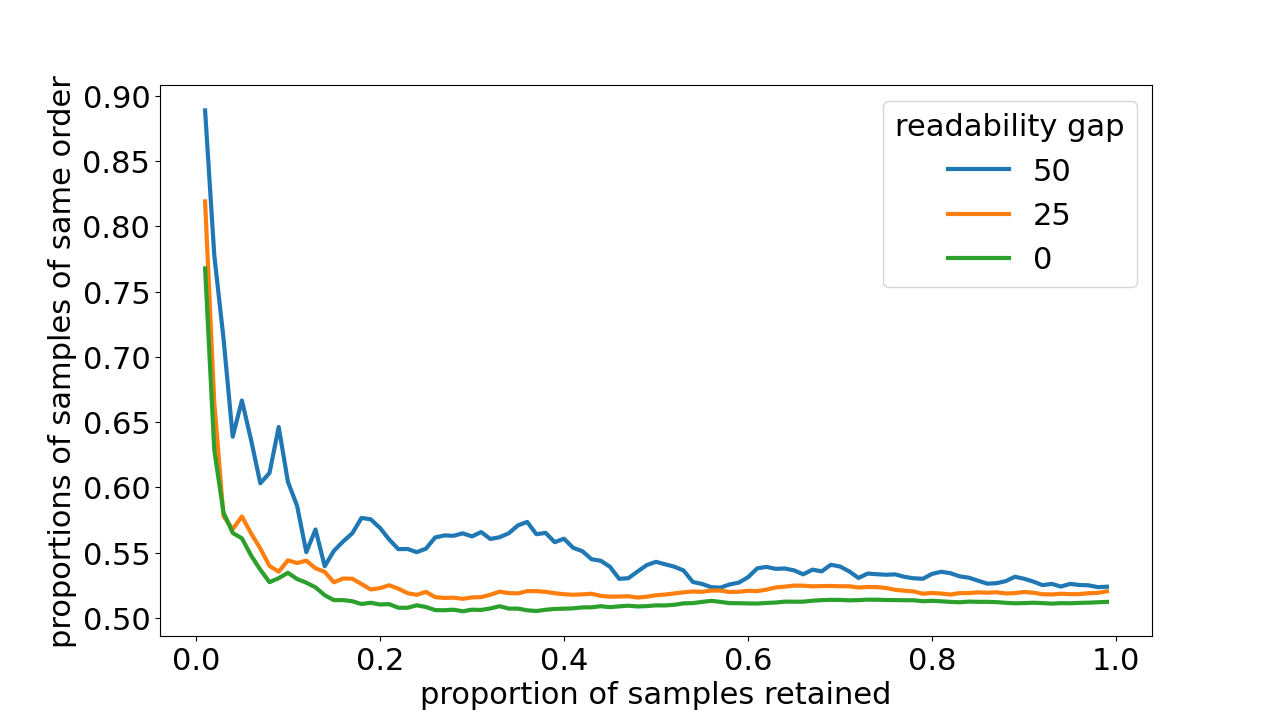}
        \caption{Reading comprehension.}
        \label{fig:sweep_samp_rc}
    \end{subfigure}%
    ~ 
    \begin{subfigure}[t]{1\columnwidth}
        \centering
\includegraphics[width=3in]{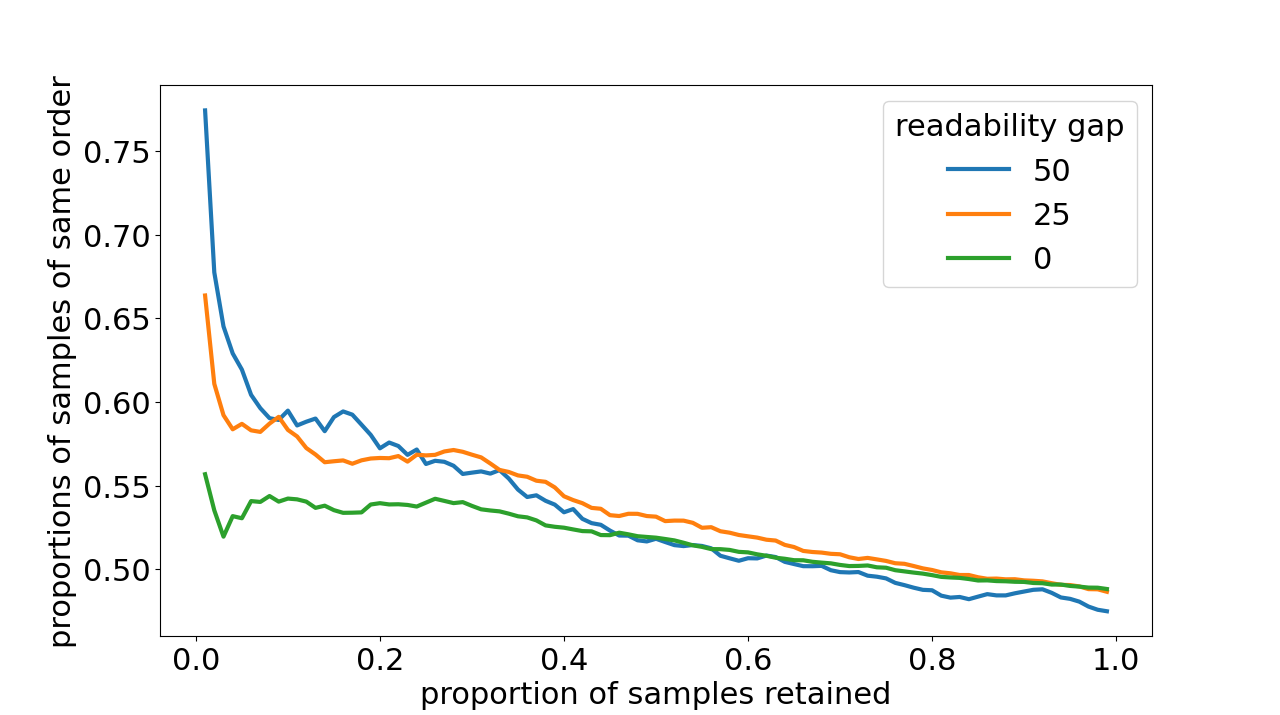}
        \caption{Sentiment classification}
        \label{fig:sweep_samp_sc}
    \end{subfigure}%
    \caption{Entropy filtered pairwise agreement in paraphrase readability and true class probability ordering with various minimum readability gaps.}
    \label{fig:entropy_sweep}
\end{figure*}

\subsection{Sentiment classification}

The inherent nature of the sentiment classification task means there is only a single input element. Therefore, we focus on exploring the relative contributions of the semantic and linguistic components of the input for this task.

Intuitively, the sentiment of a passage of text should be determined solely by its semantic content. However, Table \ref{tab:test_accs_sent} shows that for three popular sentiment classification datasets, the linguistic realization does play a role on the returned output distribution. In particular, the relative linguistic component for Amazon hits 10\% of the total influence. 

It appears the semantic component is more dominant for IMDb and Yelp compared to Amazon. One possible reason is that the length of texts is shorter for Amazon compared to the other datasets (see the data statistics in Table \ref{tab:test_data}). Hence, longer texts have a greater opportunity to reinforce sentiment being expressed, which makes it more robust to different linguistic realizations. Appendix \ref{app:res-sc} further explores this hypothesis by considering additional sentiment classification datasets.



\subsection{Impact of linguistic realization}

Section \ref{sec:res-rc} demonstrated that the linguistic realization of the context in multiple-choice reading comprehension has measurable influence on the output distribution over the options. The different linguistic realizations in this work were generated by paraphrasing the context at a range of text readability levels (see Section \ref{sec:linguistic}). This ablation study investigates the relationship of the influence of the linguistic realization i.e. the correlation between the readability of a given paraphrase of the context and the output probability of the true class (termed true class probability henceforth).

In order to determine the relationship between readability and true class probability, an entropy filter is applied to remove examples for which the entropy of the output distribution is too high, as high entropy examples suggest a random guess and hence challenging to ascertain whether a correlation exists.
Figure \ref{fig:sweep_samp_rc} sweeps the fraction of examples retained according to entropy filter and plots the fraction of the remaining examples for which the rankings of true class probability scores for every pair of paraphrases (of a given context) matches their real readability scores. The plots are indicated for minimum readability gaps of 0, 25 and 50 for the pairs of paraphrases. It is observed that the readability of a paraphrase corresponds to the returned true class probability, with a stronger correlation when the minimum readability gap between the pairs of paraphrases is higher.

A similar process is applied for the task of sentiment classification in Figure \ref{fig:sweep_samp_sc} to determine the relationship between the readability of the linguistic realization of a context and the true class probability. Like multiple choice reading comprehension, Figure \ref{fig:sweep_samp_sc} shows that there is a positive correlation between the readability level and the true class probability, with a more pronounced relationship by constraining the pairs of paraphrases to have a larger readability gap.

\section{Conclusions}

This work describes a general information-theoretic framework for text classification tasks. The framework devises a strategy for determining the influence of each input element on the final output. Additionally, each input element is further partitioned into its semantic and linguistic components. Multiple-choice reading comprehension and sentiment classification are considered as case study tasks for analysis.

For multiple-choice reading comprehension, it is found that both the context and question elements play influential roles on the final output distribution. It is further established that selection of more challenging contexts (more difficult to read) permits a greater variation (in terms of complexity) of questions to be asked about the context. Simpler contexts limit the range of the complexity to only easy questions. Hence, content creators need to carefully consider the choice of the context passage when designing multiple-choice questions to cater to a range of difficulty levels. 

In sentiment classification, it is found that the linguistic realization of the input has a measurable impact on the output distribution. Hence, the wording of the text cannot be neglected when deducing the sentiment.
For both tasks, it is determined that greater the readability (higher readability indicates the text is easier to read) of a specific linguistic realization of the text, easier the example (i.e. greater probability mass is attributed to the true class in the output distribution).

\section{Future work}

The analysis in this work has applied the framework to specifically NLP classification tasks. It would be interesting to extend the framework to both regression and sequence output tasks. For sequential outputs, there needs to be a methodology to convert the generated sequence to a single score such that its sensitivity can be measured to each input element. 

The framework applied to textual data to explore the influence of semantic vs linguistic components can also be extended to image inputs. Here, we can perceive the semantic content as the object being described in the image while the linguistic realization is based on the recording equipment that controls aspects such as orientation, resolution (blurring), camera angle, e.t.c. Therefore, the proposed information-theoretic approach has potential applications across several modalities.


\section{Limitations}

This work has several assumptions that must be stated. For the multiple-choice reading comprehension analysis, the question influence is based on real questions generated by humans on the original context. However, there is the possibility that the set of questions on a given context are not generated independently but instead the question creator has curated the question set together. Additionally, it is assumed that the paraphrasing of texts only changes the linguistic realization. However, it is likely that it also has an impact on the semantic content to an extent, which is reflected in the linguistic component influence on the output.

\section{Acknowledgements}
This research is partially funded by the EPSRC (The Engineering
and Physical Sciences Research Council)
Doctoral Training Partnership (DTP) PhD studentship
and supported by Cambridge University Press \& Assessment (CUP\&A), a
department of The Chancellor, Masters, and Scholars
of the University of Cambridge.

\newpage


\bibliographystyle{acl_natbib}
\bibliography{import}

\begin{thebibliography}{72}
\expandafter\ifx\csname natexlab\endcsname\relax\def\natexlab#1{#1}\fi

\bibitem[{AlShourbaji et~al.(2023)AlShourbaji, Helian, Sun, Hussien, Abualigah, and Elnaim}]{alshourbaji2023efficient}
Ibrahim AlShourbaji, Na~Helian, Yi~Sun, Abdelazim~G Hussien, Laith Abualigah, and Bushra Elnaim. 2023.
\newblock An efficient churn prediction model using gradient boosting machine and metaheuristic optimization.
\newblock \emph{Scientific Reports}, 13(1):14441.

\bibitem[{Banerjee and Lavie(2005)}]{banerjee2005meteor}
Satanjeev Banerjee and Alon Lavie. 2005.
\newblock Meteor: An automatic metric for mt evaluation with improved correlation with human judgments.
\newblock In \emph{Proceedings of the acl workshop on intrinsic and extrinsic evaluation measures for machine translation and/or summarization}, pages 65--72.

\bibitem[{Baradaran et~al.(2022)Baradaran, Ghiasi, and Amirkhani}]{baradaran2022survey}
Razieh Baradaran, Razieh Ghiasi, and Hossein Amirkhani. 2022.
\newblock A survey on machine reading comprehension systems.
\newblock \emph{Natural Language Engineering}, 28(6):683--732.

\bibitem[{Barbieri et~al.(2020)Barbieri, Camacho-Collados, Anke, and Neves}]{barbieri2020tweeteval}
Francesco Barbieri, Jose Camacho-Collados, Luis~Espinosa Anke, and Leonardo Neves. 2020.
\newblock Tweeteval: Unified benchmark and comparative evaluation for tweet classification.
\newblock In \emph{Findings of the Association for Computational Linguistics: EMNLP 2020}, pages 1644--1650.

\bibitem[{Beltagy et~al.(2020)Beltagy, Peters, and Cohan}]{beltagy2020longformer}
Iz~Beltagy, Matthew~E Peters, and Arman Cohan. 2020.
\newblock Longformer: The long-document transformer.
\newblock \emph{arXiv preprint arXiv:2004.05150}.

\bibitem[{Benedetto(2023)}]{benedetto2023quantitative}
Luca Benedetto. 2023.
\newblock A quantitative study of nlp approaches to question difficulty estimation.
\newblock \emph{arXiv preprint arXiv:2305.10236}.

\bibitem[{Chew et~al.(2023)Chew, Huang, Chang, and Lin}]{chew2023understanding}
Oscar Chew, Kuan-Hao Huang, Kai-Wei Chang, and Hsuan-Tien Lin. 2023.
\newblock Understanding and mitigating spurious correlations in text classification.
\newblock \emph{arXiv preprint arXiv:2305.13654}.

\bibitem[{Chowdhary and Chowdhary(2020)}]{chowdhary2020natural}
KR1442 Chowdhary and KR~Chowdhary. 2020.
\newblock Natural language processing.
\newblock \emph{Fundamentals of artificial intelligence}, pages 603--649.

\bibitem[{Clark et~al.(2020)Clark, Luong, Le, and Manning}]{clark2020electra}
Kevin Clark, Minh-Thang Luong, Quoc~V Le, and Christopher~D Manning. 2020.
\newblock Electra: Pre-training text encoders as discriminators rather than generators.
\newblock \emph{arXiv preprint arXiv:2003.10555}.

\bibitem[{Costa and Pedreira(2023)}]{costa2023recent}
Vin{\'\i}cius~G Costa and Carlos~E Pedreira. 2023.
\newblock Recent advances in decision trees: An updated survey.
\newblock \emph{Artificial Intelligence Review}, 56(5):4765--4800.

\bibitem[{Depeweg et~al.(2018)Depeweg, Hernandez-Lobato, Doshi-Velez, and Udluft}]{depeweg2018decomposition}
Stefan Depeweg, Jose-Miguel Hernandez-Lobato, Finale Doshi-Velez, and Steffen Udluft. 2018.
\newblock Decomposition of uncertainty in bayesian deep learning for efficient and risk-sensitive learning.
\newblock In \emph{International Conference on Machine Learning}, pages 1184--1193. PMLR.

\bibitem[{Dettmers et~al.(2023)Dettmers, Pagnoni, Holtzman, and Zettlemoyer}]{dettmers2023qlora}
Tim Dettmers, Artidoro Pagnoni, Ari Holtzman, and Luke Zettlemoyer. 2023.
\newblock Qlora: Efficient finetuning of quantized llms.
\newblock \emph{arXiv preprint arXiv:2305.14314}.

\bibitem[{Devlin et~al.(2018)Devlin, Chang, Lee, and Toutanova}]{devlin2018bert}
Jacob Devlin, Ming-Wei Chang, Kenton Lee, and Kristina Toutanova. 2018.
\newblock Bert: Pre-training of deep bidirectional transformers for language understanding.
\newblock \emph{arXiv preprint arXiv:1810.04805}.

\bibitem[{Dua et~al.(2019)Dua, Wang, Dasigi, Stanovsky, Singh, and Gardner}]{dua2019drop}
Dheeru Dua, Yizhong Wang, Pradeep Dasigi, Gabriel Stanovsky, Sameer Singh, and Matt Gardner. 2019.
\newblock Drop: A reading comprehension benchmark requiring discrete reasoning over paragraphs.
\newblock \emph{arXiv preprint arXiv:1903.00161}.

\bibitem[{Dugan et~al.(2022)Dugan, Miltsakaki, Upadhyay, Ginsberg, Gonzalez, Choi, Yuan, and Callison-Burch}]{dugan2022feasibility}
Liam Dugan, Eleni Miltsakaki, Shriyash Upadhyay, Etan Ginsberg, Hannah Gonzalez, Dayheon Choi, Chuning Yuan, and Chris Callison-Burch. 2022.
\newblock A feasibility study of answer-agnostic question generation for education.
\newblock \emph{arXiv preprint arXiv:2203.08685}.

\bibitem[{Fan and Lv(2008)}]{fan2008sure}
Jianqing Fan and Jinchi Lv. 2008.
\newblock Sure independence screening for ultrahigh dimensional feature space.
\newblock \emph{Journal of the Royal Statistical Society Series B: Statistical Methodology}, 70(5):849--911.

\bibitem[{Farajidizaji et~al.(2023)Farajidizaji, Raina, and Gales}]{farajidizaji2023possible}
Asma Farajidizaji, Vatsal Raina, and Mark Gales. 2023.
\newblock Is it possible to modify text to a target readability level? an initial investigation using zero-shot large language models.
\newblock \emph{arXiv preprint arXiv:2309.12551}.

\bibitem[{Flesch(1948)}]{flesch1948new}
Rudolph Flesch. 1948.
\newblock A new readability yardstick.
\newblock \emph{Journal of applied psychology}, 32(3):221.

\bibitem[{FRIZELLE et~al.(2017)FRIZELLE, O'NEILL, and BISHOP}]{frizelle_o'neill_bishop_2017}
PAULINE FRIZELLE, CLODAGH O'NEILL, and DOROTHY V.~M. BISHOP. 2017.
\newblock \href {https://doi.org/10.1017/S0305000916000635} {Assessing understanding of relative clauses: a comparison of multiple-choice comprehension versus sentence repetition}.
\newblock \emph{Journal of Child Language}, 44(6):1435–1457.

\bibitem[{Gao et~al.(2018)Gao, Bing, Chen, Lyu, and King}]{gao2018difficulty}
Yifan Gao, Lidong Bing, Wang Chen, Michael~R Lyu, and Irwin King. 2018.
\newblock Difficulty controllable generation of reading comprehension questions.
\newblock \emph{arXiv preprint arXiv:1807.03586}.

\bibitem[{Gao et~al.(2019)Gao, Bing, Li, King, and Lyu}]{gao2019generating}
Yifan Gao, Lidong Bing, Piji Li, Irwin King, and Michael~R Lyu. 2019.
\newblock Generating distractors for reading comprehension questions from real examinations.
\newblock In \emph{Proceedings of the AAAI Conference on Artificial Intelligence}, volume~33, pages 6423--6430.

\bibitem[{Guo et~al.(2017)Guo, Pleiss, Sun, and Weinberger}]{guo2017calibration}
Chuan Guo, Geoff Pleiss, Yu~Sun, and Kilian~Q Weinberger. 2017.
\newblock On calibration of modern neural networks.
\newblock In \emph{International conference on machine learning}, pages 1321--1330. PMLR.

\bibitem[{Huang et~al.(2023)Huang, Das, and Tsuda}]{huang2023feature}
Chao Huang, Diptesh Das, and Koji Tsuda. 2023.
\newblock Feature importance measurement based on decision tree sampling.
\newblock \emph{arXiv preprint arXiv:2307.13333}.

\bibitem[{Hwang and Song(2023)}]{hwang2023recent}
Yejin Hwang and Jongwoo Song. 2023.
\newblock Recent deep learning methods for tabular data.
\newblock \emph{Communications for Statistical Applications and Methods}, 30(2):215--226.

\bibitem[{Jiang et~al.(2023)Jiang, Sablayrolles, Mensch, Bamford, Chaplot, Casas, Bressand, Lengyel, Lample, Saulnier et~al.}]{jiang2023mistral}
Albert~Q Jiang, Alexandre Sablayrolles, Arthur Mensch, Chris Bamford, Devendra~Singh Chaplot, Diego de~las Casas, Florian Bressand, Gianna Lengyel, Guillaume Lample, Lucile Saulnier, et~al. 2023.
\newblock Mistral 7b.
\newblock \emph{arXiv preprint arXiv:2310.06825}.

\bibitem[{Jiang et~al.(2024)Jiang, Sablayrolles, Roux, Mensch, Savary, Bamford, Chaplot, Casas, Hanna, Bressand et~al.}]{jiang2024mixtral}
Albert~Q Jiang, Alexandre Sablayrolles, Antoine Roux, Arthur Mensch, Blanche Savary, Chris Bamford, Devendra~Singh Chaplot, Diego de~las Casas, Emma~Bou Hanna, Florian Bressand, et~al. 2024.
\newblock Mixtral of experts.
\newblock \emph{arXiv preprint arXiv:2401.04088}.

\bibitem[{Khashabi et~al.(2018)Khashabi, Chaturvedi, Roth, Upadhyay, and Roth}]{khashabi2018looking}
Daniel Khashabi, Snigdha Chaturvedi, Michael Roth, Shyam Upadhyay, and Dan Roth. 2018.
\newblock Looking beyond the surface: A challenge set for reading comprehension over multiple sentences.
\newblock In \emph{Proceedings of the 2018 Conference of the North American Chapter of the Association for Computational Linguistics: Human Language Technologies, Volume 1 (Long Papers)}, pages 252--262.

\bibitem[{Kurdi et~al.(2020)Kurdi, Leo, Parsia, Sattler, and Al-Emari}]{kurdi2020systematic}
Ghader Kurdi, Jared Leo, Bijan Parsia, Uli Sattler, and Salam Al-Emari. 2020.
\newblock A systematic review of automatic question generation for educational purposes.
\newblock \emph{International Journal of Artificial Intelligence in Education}, 30:121--204.

\bibitem[{Lai et~al.(2017{\natexlab{a}})Lai, Xie, Liu, Yang, and Hovy}]{Lai2017RACELR}
Guokun Lai, Qizhe Xie, Hanxiao Liu, Yiming Yang, and E.~Hovy. 2017{\natexlab{a}}.
\newblock Race: Large-scale reading comprehension dataset from examinations.
\newblock In \emph{EMNLP}.

\bibitem[{Lai et~al.(2017{\natexlab{b}})Lai, Xie, Liu, Yang, and Hovy}]{lai2017race}
Guokun Lai, Qizhe Xie, Hanxiao Liu, Yiming Yang, and Eduard Hovy. 2017{\natexlab{b}}.
\newblock Race: Large-scale reading comprehension dataset from examinations.
\newblock \emph{arXiv preprint arXiv:1704.04683}.

\bibitem[{Levesque et~al.(2012)Levesque, Davis, and Morgenstern}]{levesque2012winograd}
Hector Levesque, Ernest Davis, and Leora Morgenstern. 2012.
\newblock The winograd schema challenge.
\newblock In \emph{Thirteenth International Conference on the Principles of Knowledge Representation and Reasoning}. Citeseer.

\bibitem[{Liang et~al.(2019{\natexlab{a}})Liang, Li, and Yin}]{liang2019new}
Yichan Liang, Jianheng Li, and Jian Yin. 2019{\natexlab{a}}.
\newblock A new multi-choice reading comprehension dataset for curriculum learning.
\newblock In \emph{Asian Conference on Machine Learning}, pages 742--757. PMLR.

\bibitem[{Liang et~al.(2019{\natexlab{b}})Liang, Li, and Yin}]{pmlr-v101-liang19a}
Yichan Liang, Jianheng Li, and Jian Yin. 2019{\natexlab{b}}.
\newblock \href {http://proceedings.mlr.press/v101/liang19a.html} {A new multi-choice reading comprehension dataset for curriculum learning}.
\newblock In \emph{Proceedings of The Eleventh Asian Conference on Machine Learning}, volume 101 of \emph{Proceedings of Machine Learning Research}, pages 742--757, Nagoya, Japan. PMLR.

\bibitem[{Liu et~al.(2019)Liu, Ott, Goyal, Du, Joshi, Chen, Levy, Lewis, Zettlemoyer, and Stoyanov}]{liu2019roberta}
Yinhan Liu, Myle Ott, Naman Goyal, Jingfei Du, Mandar Joshi, Danqi Chen, Omer Levy, Mike Lewis, Luke Zettlemoyer, and Veselin Stoyanov. 2019.
\newblock Roberta: A robustly optimized bert pretraining approach.
\newblock \emph{arXiv preprint arXiv:1907.11692}.

\bibitem[{Liusie et~al.(2023{\natexlab{a}})Liusie, Manakul, and Gales}]{liusie2023zero}
Adian Liusie, Potsawee Manakul, and Mark~JF Gales. 2023{\natexlab{a}}.
\newblock Zero-shot nlg evaluation through pairware comparisons with llms.
\newblock \emph{arXiv preprint arXiv:2307.07889}.

\bibitem[{Liusie et~al.(2023{\natexlab{b}})Liusie, Raina, and Gales}]{liusie2023world}
Adian Liusie, Vatsal Raina, and Mark Gales. 2023{\natexlab{b}}.
\newblock " world knowledge" in multiple choice reading comprehension.
\newblock In \emph{The Sixth Fact Extraction and VERification Workshop}, page~49.

\bibitem[{Liusie et~al.(2023{\natexlab{c}})Liusie, Raina, Mullooly, Knill, and Gales}]{liusie2023analysis}
Adian Liusie, Vatsal Raina, Andrew Mullooly, Kate Knill, and Mark J.~F. Gales. 2023{\natexlab{c}}.
\newblock \href {http://arxiv.org/abs/2306.13047} {Analysis of the cambridge multiple-choice questions reading dataset with a focus on candidate response distribution}.

\bibitem[{Liusie et~al.(2022)Liusie, Raina, Raina, and Gales}]{liusie2022analyzing}
Adian Liusie, Vatsal Raina, Vyas Raina, and Mark Gales. 2022.
\newblock Analyzing biases to spurious correlations in text classification tasks.
\newblock In \emph{Proceedings of the 2nd Conference of the Asia-Pacific Chapter of the Association for Computational Linguistics and the 12th International Joint Conference on Natural Language Processing}, pages 78--84.

\bibitem[{Maas et~al.(2011)Maas, Daly, Pham, Huang, Ng, and Potts}]{maas2011learning}
Andrew Maas, Raymond~E Daly, Peter~T Pham, Dan Huang, Andrew~Y Ng, and Christopher Potts. 2011.
\newblock Learning word vectors for sentiment analysis.
\newblock In \emph{Proceedings of the 49th annual meeting of the association for computational linguistics: Human language technologies}, pages 142--150.

\bibitem[{Malinin et~al.(2021)Malinin, Band, Chesnokov, Gal, Gales, Noskov, Ploskonosov, Prokhorenkova, Provilkov, Raina et~al.}]{malinin2021shifts}
Andrey Malinin, Neil Band, German Chesnokov, Yarin Gal, Mark~JF Gales, Alexey Noskov, Andrey Ploskonosov, Liudmila Prokhorenkova, Ivan Provilkov, Vatsal Raina, et~al. 2021.
\newblock Shifts: A dataset of real distributional shift across multiple large-scale tasks.
\newblock \emph{arXiv preprint arXiv:2107.07455}.

\bibitem[{Malinin and Gales(2018)}]{malinin2018predictive}
Andrey Malinin and Mark Gales. 2018.
\newblock Predictive uncertainty estimation via prior networks.
\newblock \emph{Advances in neural information processing systems}, 31.

\bibitem[{Manakul et~al.(2023)Manakul, Liusie, and Gales}]{manakul2023mqag}
Potsawee Manakul, Adian Liusie, and Mark~JF Gales. 2023.
\newblock Mqag: Multiple-choice question answering and generation for assessing information consistency in summarization.
\newblock \emph{arXiv preprint arXiv:2301.12307}.

\bibitem[{McAuley and Leskovec(2013)}]{mcauley2013hidden}
Julian McAuley and Jure Leskovec. 2013.
\newblock Hidden factors and hidden topics: understanding rating dimensions with review text.
\newblock In \emph{Proceedings of the 7th ACM conference on Recommender systems}, pages 165--172.

\bibitem[{Mullooly et~al.(2023)Mullooly, Andersen, Benedetto, Buttery, Caines, Gales, Karatay, Knill, Liusie, Raina, and Taslimipoor}]{CMCQRD-2023}
Andrew Mullooly, {\O}istein Andersen, Luca Benedetto, Paula Buttery, Andrew Caines, Mark~J.F. Gales, Yasin Karatay, Kate Knill, Adian Liusie, Vatsal Raina, and Shiva Taslimipoor. 2023.
\newblock {The Cambridge Multiple-Choice Questions Reading Dataset}.
\newblock Cambridge University Press and Assessment.

\bibitem[{Ni et~al.(2022)Ni, Abrego, Constant, Ma, Hall, Cer, and Yang}]{ni2022sentence}
Jianmo Ni, Gustavo~Hernandez Abrego, Noah Constant, Ji~Ma, Keith Hall, Daniel Cer, and Yinfei Yang. 2022.
\newblock Sentence-t5: Scalable sentence encoders from pre-trained text-to-text models.
\newblock In \emph{Findings of the Association for Computational Linguistics: ACL 2022}, pages 1864--1874.

\bibitem[{Och(2003)}]{och2003minimum}
Franz~Josef Och. 2003.
\newblock Minimum error rate training in statistical machine translation.
\newblock In \emph{Proceedings of the 41st annual meeting of the Association for Computational Linguistics}, pages 160--167.

\bibitem[{Pang and Lee(2005)}]{pang-lee-2005-seeing}
Bo~Pang and Lillian Lee. 2005.
\newblock \href {https://doi.org/10.3115/1219840.1219855} {Seeing stars: Exploiting class relationships for sentiment categorization with respect to rating scales}.
\newblock In \emph{Proceedings of the 43rd Annual Meeting of the Association for Computational Linguistics ({ACL}{'}05)}, pages 115--124, Ann Arbor, Michigan. Association for Computational Linguistics.

\bibitem[{Pati et~al.(1993)Pati, Rezaiifar, and Krishnaprasad}]{pati1993orthogonal}
Yagyensh~Chandra Pati, Ramin Rezaiifar, and Perinkulam~Sambamurthy Krishnaprasad. 1993.
\newblock Orthogonal matching pursuit: Recursive function approximation with applications to wavelet decomposition.
\newblock In \emph{Proceedings of 27th Asilomar conference on signals, systems and computers}, pages 40--44. IEEE.

\bibitem[{Raina and Gales(2022{\natexlab{a}})}]{raina-gales-2022-answer}
Vatsal Raina and Mark Gales. 2022{\natexlab{a}}.
\newblock \href {https://doi.org/10.18653/v1/2022.findings-acl.82} {Answer uncertainty and unanswerability in multiple-choice machine reading comprehension}.
\newblock In \emph{Findings of the Association for Computational Linguistics: ACL 2022}, pages 1020--1034, Dublin, Ireland. Association for Computational Linguistics.

\bibitem[{Raina and Gales(2022{\natexlab{b}})}]{raina2022multiple}
Vatsal Raina and Mark Gales. 2022{\natexlab{b}}.
\newblock Multiple-choice question generation: Towards an automated assessment framework.
\newblock \emph{arXiv preprint arXiv:2209.11830}.

\bibitem[{Raina et~al.(2023{\natexlab{a}})Raina, Kassner, Popat, Lewis, Cancedda, and Martin}]{raina-etal-2023-erate}
Vatsal Raina, Nora Kassner, Kashyap Popat, Patrick Lewis, Nicola Cancedda, and Louis Martin. 2023{\natexlab{a}}.
\newblock \href {https://doi.org/10.18653/v1/2023.insights-1.2} {{ERATE}: Efficient retrieval augmented text embeddings}.
\newblock In \emph{The Fourth Workshop on Insights from Negative Results in NLP}, pages 11--18, Dubrovnik, Croatia. Association for Computational Linguistics.

\bibitem[{Raina et~al.(2023{\natexlab{b}})Raina, Liusie, and Gales}]{raina2023analyzing}
Vatsal Raina, Adian Liusie, and Mark Gales. 2023{\natexlab{b}}.
\newblock Analyzing multiple-choice reading and listening comprehension tests.
\newblock \emph{arXiv preprint arXiv:2307.01076}.

\bibitem[{Richardson et~al.(2013{\natexlab{a}})Richardson, Burges, and Renshaw}]{richardson-etal-2013-mctest}
Matthew Richardson, Christopher~J.C. Burges, and Erin Renshaw. 2013{\natexlab{a}}.
\newblock \href {https://aclanthology.org/D13-1020} {{MCT}est: A challenge dataset for the open-domain machine comprehension of text}.
\newblock In \emph{Proceedings of the 2013 Conference on Empirical Methods in Natural Language Processing}, pages 193--203, Seattle, Washington, USA. Association for Computational Linguistics.

\bibitem[{Richardson et~al.(2013{\natexlab{b}})Richardson, Burges, and Renshaw}]{richardson2013mctest}
Matthew Richardson, Christopher~JC Burges, and Erin Renshaw. 2013{\natexlab{b}}.
\newblock Mctest: A challenge dataset for the open-domain machine comprehension of text.
\newblock In \emph{Proceedings of the 2013 conference on empirical methods in natural language processing}, pages 193--203.

\bibitem[{Rudin(2019)}]{rudin2019stop}
Cynthia Rudin. 2019.
\newblock Stop explaining black box machine learning models for high stakes decisions and use interpretable models instead.
\newblock \emph{Nature machine intelligence}, 1(5):206--215.

\bibitem[{Sanh et~al.(2019)Sanh, Debut, Chaumond, and Wolf}]{sanh2019distilbert}
Victor Sanh, Lysandre Debut, Julien Chaumond, and Thomas Wolf. 2019.
\newblock Distilbert, a distilled version of bert: Smaller, faster, cheaper and lighter. arxiv 2019.
\newblock \emph{arXiv preprint arXiv:1910.01108}.

\bibitem[{Socher et~al.(2013)Socher, Perelygin, Wu, Chuang, Manning, Ng, and Potts}]{socher-etal-2013-recursive}
Richard Socher, Alex Perelygin, Jean Wu, Jason Chuang, Christopher~D. Manning, Andrew Ng, and Christopher Potts. 2013.
\newblock \href {https://www.aclweb.org/anthology/D13-1170} {Recursive deep models for semantic compositionality over a sentiment treebank}.
\newblock In \emph{Proceedings of the 2013 Conference on Empirical Methods in Natural Language Processing}, pages 1631--1642, Seattle, Washington, USA. Association for Computational Linguistics.

\bibitem[{Stahlberg(2020)}]{stahlberg2020neural}
Felix Stahlberg. 2020.
\newblock Neural machine translation: A review.
\newblock \emph{Journal of Artificial Intelligence Research}, 69:343--418.

\bibitem[{Sugawara et~al.(2022)Sugawara, Nangia, Warstadt, and Bowman}]{sugawara2022makes}
Saku Sugawara, Nikita Nangia, Alex Warstadt, and Samuel~R Bowman. 2022.
\newblock What makes reading comprehension questions difficult?
\newblock \emph{arXiv preprint arXiv:2203.06342}.

\bibitem[{Sun et~al.(2019)Sun, Yu, Chen, Yu, Choi, and Cardie}]{sun2019dream}
Kai Sun, Dian Yu, Jianshu Chen, Dong Yu, Yejin Choi, and Claire Cardie. 2019.
\newblock Dream: A challenge data set and models for dialogue-based reading comprehension.
\newblock \emph{Transactions of the Association for Computational Linguistics}, 7:217--231.

\bibitem[{Tibshirani(1996)}]{tibshirani1996regression}
Robert Tibshirani. 1996.
\newblock Regression shrinkage and selection via the lasso.
\newblock \emph{Journal of the Royal Statistical Society Series B: Statistical Methodology}, 58(1):267--288.

\bibitem[{Touvron et~al.(2023{\natexlab{a}})Touvron, Lavril, Izacard, Martinet, Lachaux, Lacroix, Rozi{\`e}re, Goyal, Hambro, Azhar et~al.}]{touvron2023llama}
Hugo Touvron, Thibaut Lavril, Gautier Izacard, Xavier Martinet, Marie-Anne Lachaux, Timoth{\'e}e Lacroix, Baptiste Rozi{\`e}re, Naman Goyal, Eric Hambro, Faisal Azhar, et~al. 2023{\natexlab{a}}.
\newblock Llama: Open and efficient foundation language models.
\newblock \emph{arXiv preprint arXiv:2302.13971}.

\bibitem[{Touvron et~al.(2023{\natexlab{b}})Touvron, Martin, Stone, Albert, Almahairi, Babaei, Bashlykov, Batra, Bhargava, Bhosale et~al.}]{touvron2023llama2}
Hugo Touvron, Louis Martin, Kevin Stone, Peter Albert, Amjad Almahairi, Yasmine Babaei, Nikolay Bashlykov, Soumya Batra, Prajjwal Bhargava, Shruti Bhosale, et~al. 2023{\natexlab{b}}.
\newblock Llama 2: Open foundation and fine-tuned chat models.
\newblock \emph{arXiv preprint arXiv:2307.09288}.

\bibitem[{Tunstall et~al.(2023)Tunstall, Beeching, Lambert, Rajani, Rasul, Belkada, Huang, von Werra, Fourrier, Habib et~al.}]{tunstall2023zephyr}
Lewis Tunstall, Edward Beeching, Nathan Lambert, Nazneen Rajani, Kashif Rasul, Younes Belkada, Shengyi Huang, Leandro von Werra, Cl{\'e}mentine Fourrier, Nathan Habib, et~al. 2023.
\newblock Zephyr: Direct distillation of lm alignment.
\newblock \emph{arXiv preprint arXiv:2310.16944}.

\bibitem[{Vaswani et~al.(2017)Vaswani, Shazeer, Parmar, Uszkoreit, Jones, Gomez, Kaiser, and Polosukhin}]{vaswani2017attention}
Ashish Vaswani, Noam Shazeer, Niki Parmar, Jakob Uszkoreit, Llion Jones, Aidan~N Gomez, {\L}ukasz Kaiser, and Illia Polosukhin. 2017.
\newblock Attention is all you need.
\newblock \emph{Advances in neural information processing systems}, 30.

\bibitem[{Wankhade et~al.(2022)Wankhade, Rao, and Kulkarni}]{wankhade2022survey}
Mayur Wankhade, Annavarapu Chandra~Sekhara Rao, and Chaitanya Kulkarni. 2022.
\newblock A survey on sentiment analysis methods, applications, and challenges.
\newblock \emph{Artificial Intelligence Review}, 55(7):5731--5780.

\bibitem[{Widyassari et~al.(2022)Widyassari, Rustad, Shidik, Noersasongko, Syukur, Affandy et~al.}]{widyassari2022review}
Adhika~Pramita Widyassari, Supriadi Rustad, Guruh~Fajar Shidik, Edi Noersasongko, Abdul Syukur, Affandy Affandy, et~al. 2022.
\newblock Review of automatic text summarization techniques \& methods.
\newblock \emph{Journal of King Saud University-Computer and Information Sciences}, 34(4):1029--1046.

\bibitem[{Xu et~al.(2023)Xu, Wang, Liao, and Wang}]{xu2023efficient}
Biao Xu, Yao Wang, Xiuwu Liao, and Kaidong Wang. 2023.
\newblock Efficient fraud detection using deep boosting decision trees.
\newblock \emph{Decision Support Systems}, page 114037.

\bibitem[{Yang et~al.(2018)Yang, Qi, Zhang, Bengio, Cohen, Salakhutdinov, and Manning}]{yang2018hotpotqa}
Zhilin Yang, Peng Qi, Saizheng Zhang, Yoshua Bengio, William~W Cohen, Ruslan Salakhutdinov, and Christopher~D Manning. 2018.
\newblock Hotpotqa: A dataset for diverse, explainable multi-hop question answering.
\newblock \emph{arXiv preprint arXiv:1809.09600}.

\bibitem[{Yu et~al.(2020)Yu, Jiang, Dong, and Feng}]{Yu2020ReClorAR}
Weihao Yu, Zihang Jiang, Yanfei Dong, and Jiashi Feng. 2020.
\newblock \href {https://openreview.net/forum?id=HJgJtT4tvB} {Reclor: A reading comprehension dataset requiring logical reasoning}.
\newblock In \emph{International Conference on Learning Representations}.

\bibitem[{Zhang et~al.(2019)Zhang, Kishore, Wu, Weinberger, and Artzi}]{zhang2019bertscore}
Tianyi Zhang, Varsha Kishore, Felix Wu, Kilian~Q Weinberger, and Yoav Artzi. 2019.
\newblock Bertscore: Evaluating text generation with bert.
\newblock \emph{arXiv preprint arXiv:1904.09675}.

\bibitem[{Zhang et~al.(2015)Zhang, Zhao, and LeCun}]{zhang2015character}
Xiang Zhang, Junbo Zhao, and Yann LeCun. 2015.
\newblock Character-level convolutional networks for text classification.
\newblock \emph{Advances in neural information processing systems}, 28.

\end{thebibliography}

\appendix
\newpage
.

\newpage

\section{Data}
\label{app:data}

Table \ref{tab:app_data_stats} provides an overview of the RACE++ dataset, which is divided into three subsets: RACE-M, RACE-H, and RACE-C. These subsets correspond to English exam materials from Chinese middle schools (RACE-M), high schools (RACE-H), and colleges (RACE-C) respectively. Table \ref{tab:app_data_stats} displays key statistics for each of these subsets including the number of contexts, the average number of questions for one context, and the semantic and linguistic diversity. 

SST-2 \citep{socher-etal-2013-recursive} and TweetEval \citep{barbieri2020tweeteval} are considered as additional datasets for sentiment classification, which were not presented in the main text. 
SST-2 (Stanford Sentiment Treebank) corpus consists of movie reviews provided by \citet{pang-lee-2005-seeing} which are classified as either positive or negative. TweetEval consists of seven heterogeneous tasks based on tweets from Twitter. Here, the focus is on the tweet-emotion task where each tweet is classified as joy, sadness, optimism or anger.
These two datasets are included here as they have shorter inputs texts than IMDb, Yelp and Amazon.

\begin{table*}[htbp!]
\small
    \centering
    \begin{tabular}{ll|cccccc}
        \toprule
& & \# examples & \# options & \# words &  \# questions & semantic diversity & linguistic diversity \\
\midrule
\multirow{3}{*}{RC} & RACE-M & 362 & 4 & 200 & 4 & $\text{0.096}_{\pm{\text{0.017}}}$ & $\text{0.017}_{\pm{\text{0.007}}}$\\
& RACE-H & 510& 4 & 308& 3.3 & $\text{0.100}_{\pm{\text{0.012}}}$ & $\text{0.016}_{\pm{\text{0.006}}}$\\
& RACE-C & 135& 4 & 375& 5.2 & $\text{0.097}_{\pm{\text{0.011}}}$ & $\text{0.018}_{\pm{\text{0.006}}}$\\
\midrule
\multirow{2}{*}{SC} & SST-2 & 500 & 2  & 20  & - & $\text{0.145}_{\pm{\text{0.021}}}$ & $\text{0.087}_{\pm{\text{0.034}}}$\\
& TweetEval & 500 & 4 & 17 & - & $\text{0.149}_{\pm{\text{0.022}}}$ & $\text{0.081}_{\pm{\text{0.033}}}$\\
        \bottomrule
    \end{tabular}
    \caption{Statistics for breakdown of additional datasets including RACE-M, RACE-H, RACE-C for RACE++ in reading comprehension and SST-2, TweetEval in sentiment classification.}
    \label{tab:app_data_stats}
\end{table*}

\section{Models}
\label{app:models}

For the multiple choice reading comprehension tasks, three models are used: Llama2, RoBERTa and Longformer.

Pretrained Llama-2 is finetuned specifically on the train split of RACE++ with hyperparameter tuning on the validation split. However, it is not computationally feasible to train all the model parameters of Llama-2. Therefore, parameter efficient finetuning is used with quantized low rank adapters (QLoRA) \citep{dettmers2023qlora}. The final training parameters finetune the model with a learning rate of 1e-4, batch size of 4, lora rank of 8 with lora $\alpha$ = 16 and dropout 0.1. The model is trained for 1 epoch taking 7 hours on an NVIDIA A100 machine. For the main paper results, the Llama-2 model is selected due to its best accuracy.

RoBERTa\footnote{Available at \url{https://huggingface.co/LIAMF-USP/roberta-large-finetuned-race}} is finetuned on the train split of the RACE dataset (RACE-M and RACE-H).
The details of the specific Longformer\footnote{Available at \url{ https://huggingface.co/potsawee/longformer-large-4096-answering-race}} are detailed in \citet{manakul2023mqag}.

For the data complexity classification system, pretrained ELECTRA-base is finetuned on the RACE++ train split with the complexity class (easy, medium or hard) as the label.
The model is trained using the AdamW optimizer, a batch size of 3, learning rate of 2e-5, max number of epochs of 3 with all inputs truncated to 512 tokens. An ensemble of 3 models is trained. Training for each model takes 3 hours on an NVIDIA V100 graphical processing unit.

For the sentiment classification task, we used the models from RoBERTa and distilBERT \citep{sanh2019distilbert} family and the they are finetuned on various datasets as explained in the following.
The train split of IMDb is used to finetune RoBERTa\footnote{Available at:  \url{https://huggingface.co/wrmurray/roberta-base-finetuned-imdb}} and BERT\footnote{Available at : \url{https://huggingface.co/lvwerra/distilbert-imdb}}. Both of these models are applied on the test sets for IMDb, Yelp and Amazon.
The models we used for SST-2 and TweetEval are finetuned n their corresponding training split, namely RoBERTa-SST2\footnote{Available at: \url{rasyosef/roberta-base-finetuned-sst2}}, distilBERT-SST2\footnote{Available at: \url{https://huggingface.co/distilbert/distilbert-base-uncased-finetunedsst-2-english}}, RoBERTa-Tweet\footnote{Available at: \url{cardiffnlp/twitter-roberta-base-dec2021-emotion}}, distilBERT-Tweet\footnote{Available at: \url{https://huggingface.co/philschmid/DistilBERT-tweet-eval-emotion}}.

\section{Question filter}
\label{app:question_filter}

Based on manual observation, the RACE++ dataset has some questions that are generated from the linguistic contents of the contexts rather than from the semantic contents. As explained in Section \ref{sec:theory-mcrc}, these questions will invalidate the theoretical assumption when calculating the influence of each component because the question would be unanswerable for a generated paraphrase that does not maintain the same linguistic information. Namely, these linguistic questions are often related to their positions in the context and are always correlated with certain key words such as `in paragraph 2'. Thus, we apply a word-matching question filter to filter out all such examples, ensuring that only relevant and contextually coherent questions are retained for further processing.
We specifically filter out all questions containing the following phrases: 
`\{number\} + word/sentence/paragraph + \{number\} + refer to/mean'.


In total, for the RACE++ dataset, approximately 6.2\% questions are found to be generated from the linguistic content of the context and thus filtered out. The effects of the filter on the element influence analysis using Llama-2 is shown in Table \ref{tab:q_filter}. It is clear the measured question influence drops as expected by 1.6\%.
{
\centering
\begin{table*}[h]
\small
    \centering
    \begin{tabular}{c|cc|ccc|cc}
    \toprule
    \multirow{2}{*}{filter}  & \multicolumn{2}{c|}{accuracy} & \multicolumn{5}{c}{influence}\\
     & original & para & total & question & context  & context-semantic & context-linguistic \\
    \midrule
    No 
    &  84.2 & 81.5 & 0.284 & 0.164 (57.7\%)& 0.120 (42.3\%) & 0.135 (81.4\%)&0.031 (18.6\%)\\
      Yes 
 & 86.0& 82.9 & 0.298 &0.161 (56.1\%)&0.131 (43.9\%)&0.108 (82.5\%)& 0.023 (17.5\%)\\
    \bottomrule
    \end{tabular}
    \caption{The effect of the question filter on element influence for Llama-2 on the RACE++ test set.}
    \label{tab:q_filter}
\end{table*}
}
\section{Paraphrasing}

\label{app:paraphrasing}

{
\begin{table*}[t]
\small
    {\renewcommand{\arraystretch}{1.5}
    \begin{tabularx}{450pt}{llX}
    \toprule
    Target & Level (US) & Prompt \\
    \midrule
    5 &  Professional& Paraphrase this document for a professional. It
should be extremely difficult to read and best
understood by university graduates.\\
    20 & College graduate& Paraphrase this document for college graduate
level (US). It should be very difficult to read and
best understood by university graduates.\\
    40 & College&  Paraphrase this document for college level (US).
It should be difficult to read.\\
    55 &  10-12th grade& Paraphrase this document for 10th-12th grade
school level (US). It should be fairly difficult to
read.
\\
    65 & 8-9th grade & Paraphrase this document for 8th/9th grade
school level (US). It should be plain English
and easily understood by 13- to 15-year-old students.
\\
    75 & 7th grade& Paraphrase this document for 7th grade school
level (US). It should be fairly easy to read.\\
    85 & 6th grade& Paraphrase this document for 6th grade school
level (US). It should be easy to read a
\\
    95 & 5th grade& Paraphrase this document for 5th grade school
level (US). It should be very easy to read and
easily understood by an average 11-year old student.\\
    \bottomrule
    \end{tabularx}
    }
    \caption{Prompts to generate paraphrases with different target readability (using FRES).}
    \label{tab:para_prompts}
\end{table*}
}

\begin{figure*}[t]
    \centering
    \begin{subfigure}[t]{1\columnwidth}
        \centering
        \includegraphics[width=4in]{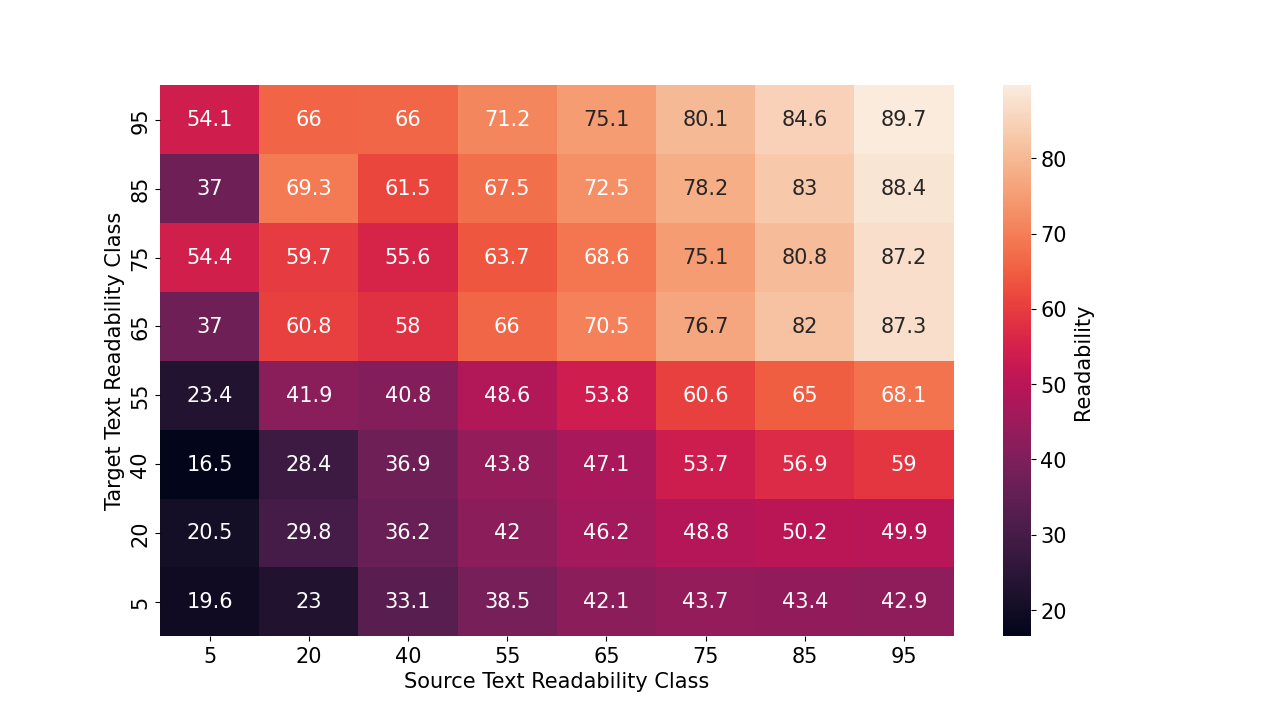}
        \caption{Reading Comprehension}
    \end{subfigure}%
    ~ 
    \begin{subfigure}[t]{1\columnwidth}
        \centering
        \includegraphics[width=4in]{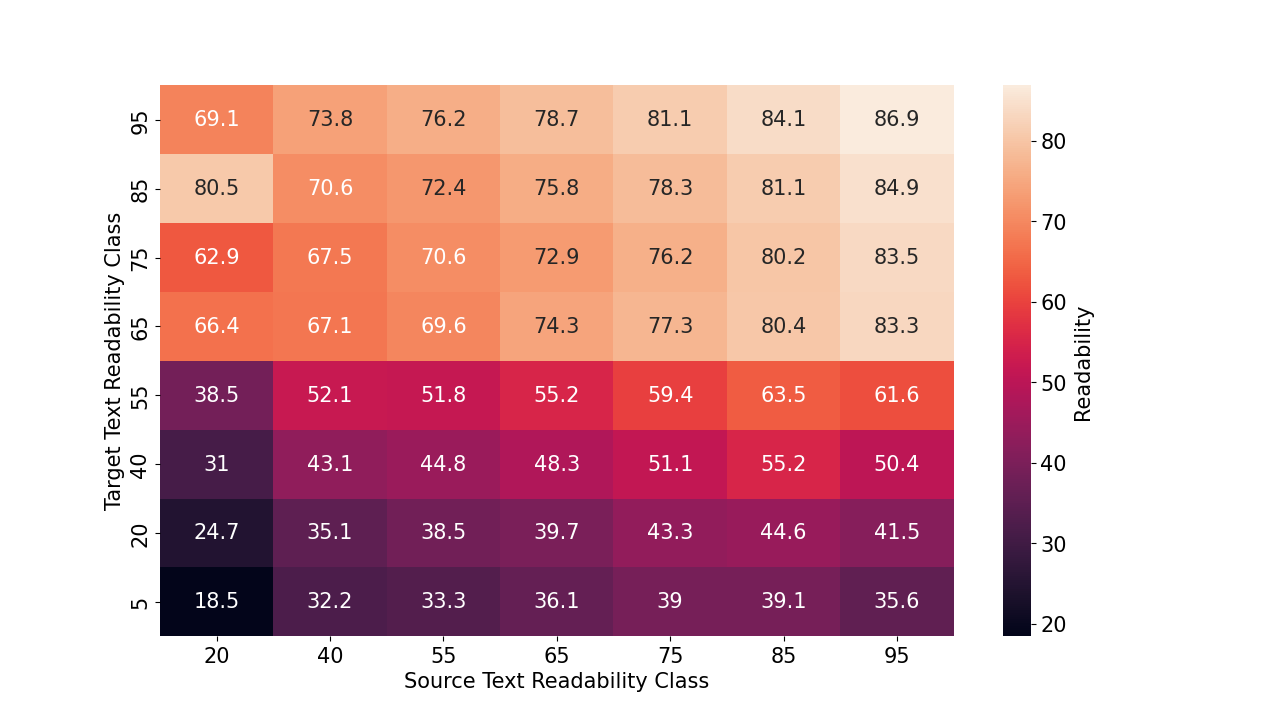}
        \caption{Sentiment Classification}
    \end{subfigure}%
    \caption{Averaged measured readability.}
    \label{fig:readability_check}
\end{figure*}

\begin{figure*}[t]
    \centering
    \begin{subfigure}[t]{1\columnwidth}
        \centering
        \includegraphics[width=4in]{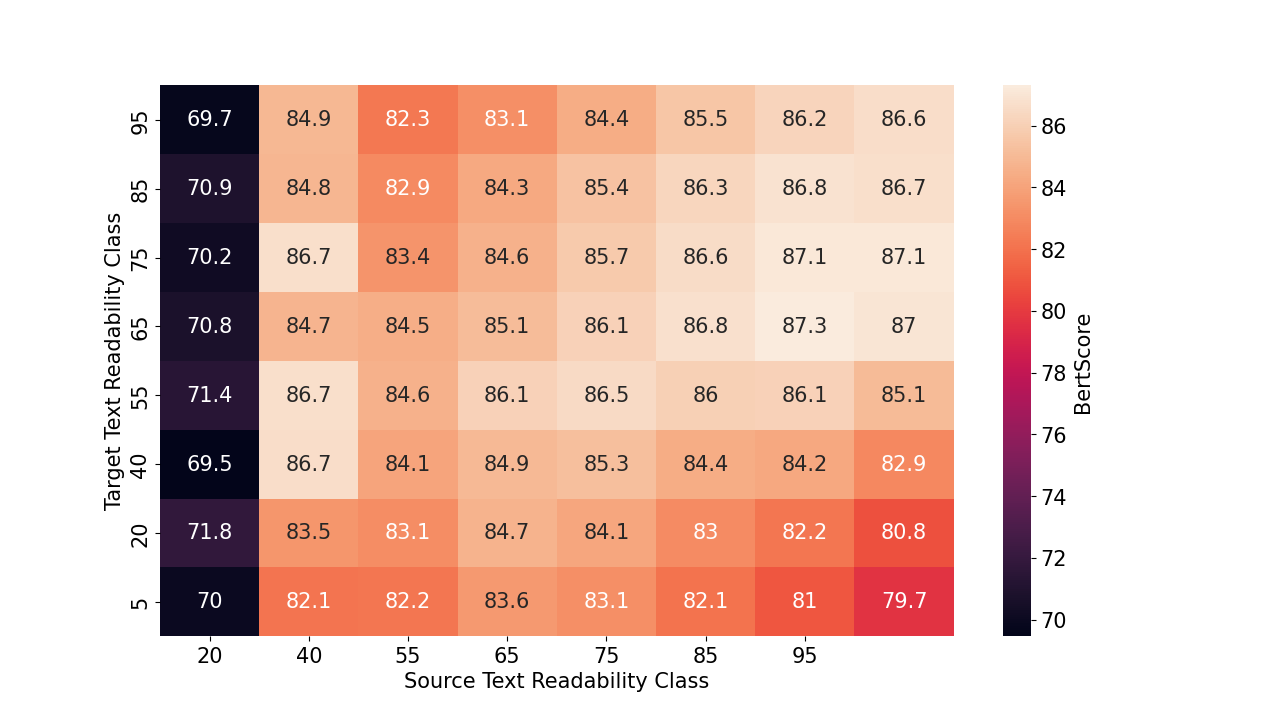}
        \caption{Reading Comprehension}
    \end{subfigure}%
    ~ 
    \begin{subfigure}[t]{1\columnwidth}
        \centering
        \includegraphics[width=4in]{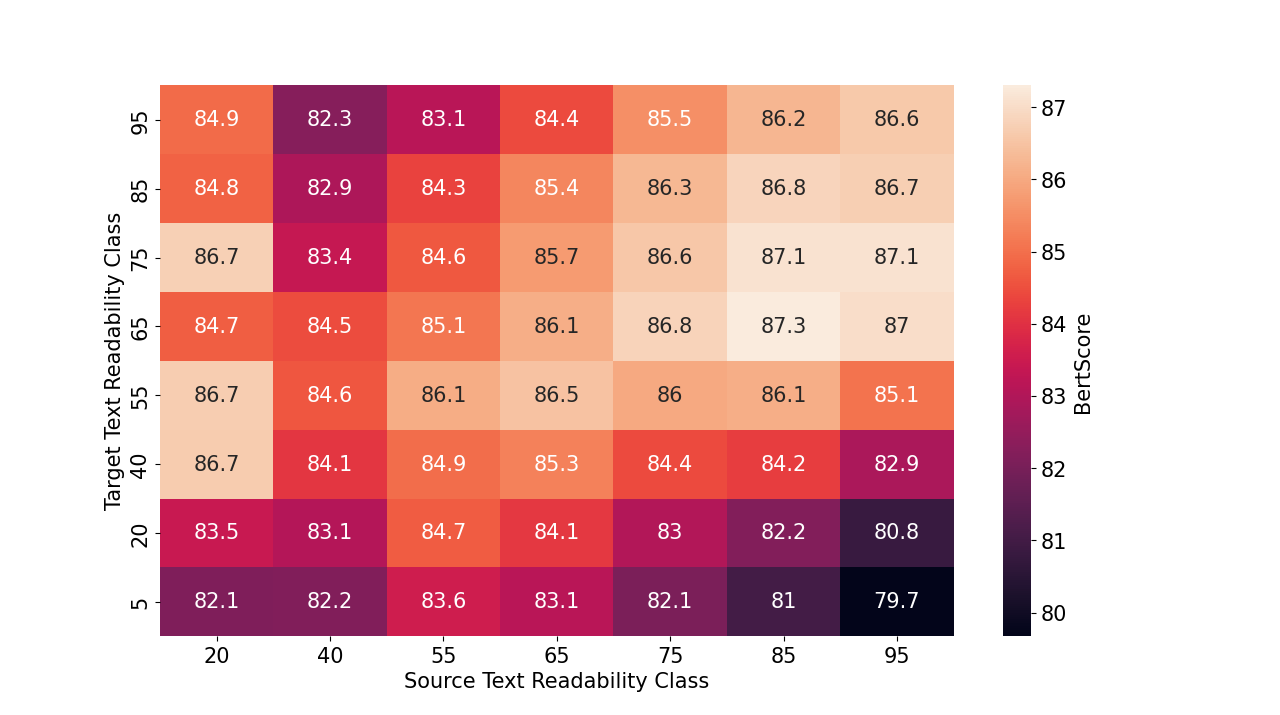}
        \caption{Sentiment Classification}
    \end{subfigure}%
    \caption{Averaged BERTScore  F1.}
    \label{fig:bertscore}
\end{figure*}

\begin{figure*}[t]
    \centering
    \begin{subfigure}[t]{1\columnwidth}
        \centering
        \includegraphics[width=4in]{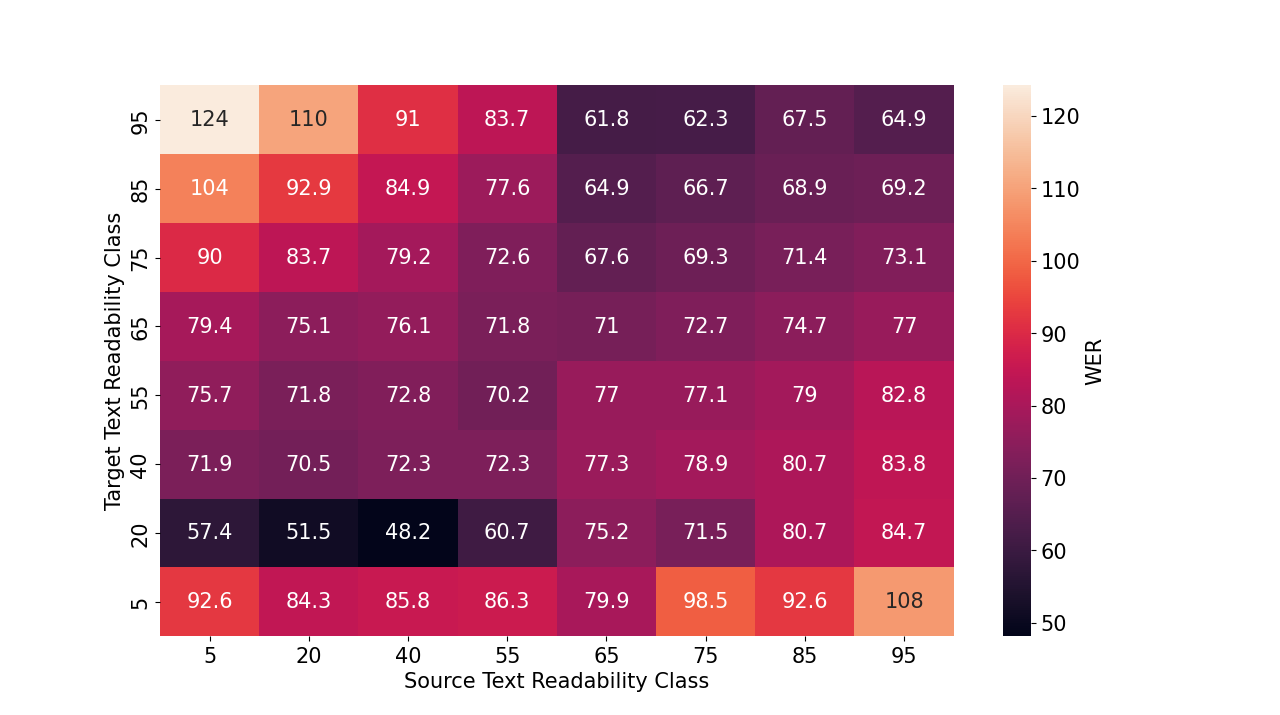}
        \caption{Reading Comprehension}
    \end{subfigure}%
    ~ 
    \begin{subfigure}[t]{1\columnwidth}
        \centering
        \includegraphics[width=4in]{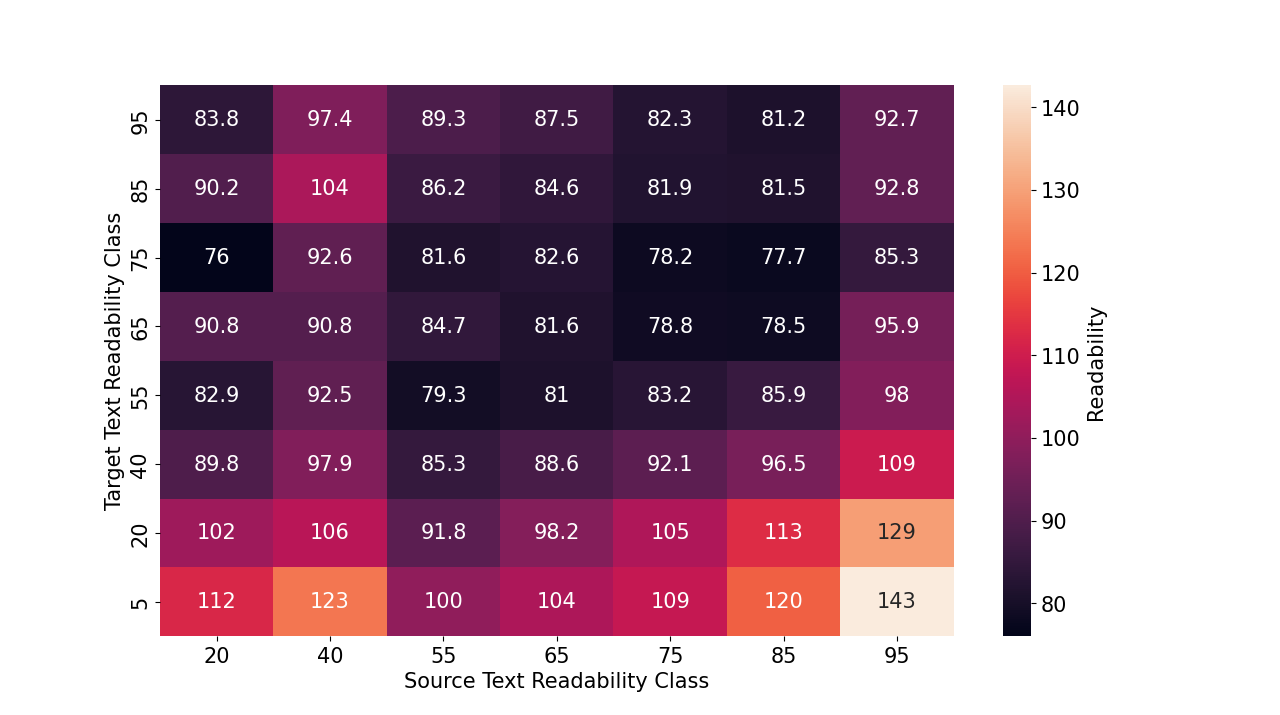}
        \caption{Sentiment Classification}
    \end{subfigure}%
    \caption{Averaged Word Error Rate.}
    \label{fig:WER}
\end{figure*}

We grouped the original texts into eight different readability levels: 5, 20, 40, 55, 65, 75, 85, and 95 for the reading comprehension task and used the final 7 groups for the sentiment classification task as there were no texts in sentiment classification that fell into the most challenging category of 0-10 on the readability scale. 
The specific prompts for each difficulty level we used are shown in Table \ref{tab:para_prompts}. Here we also present the quality of our paraphrase generation process. Figure \ref{fig:readability_check} displays the average readability score of the paraphrased text for each combination of original and target readability levels. From the heatmap, we can see that while the readability of the paraphrased text is influenced by the readability of the original text, the paraphrases still fall within an acceptable range of readability. We also report the averaged BertScore F1 \citep{zhang2019bertscore} and Word Error Rate (WER) \citep{och2003minimum} to ensure the quality of our paraphrasing system as shown in Figure\ref{fig:bertscore} and Figure \ref{fig:WER}.
An ideal paraphrasing system should expect high semantic similarity with high BERTScore and low linguistic similarity with high WER.


\section{Additional results}
Here we present the results from some additional experiments that act as a supplement to the main paper.

\subsection{Reading comprehension}

\subsubsection{Data complexity classifier}
\label{app:data_comp}

In Section \ref{sec:data_comp_model}, we used the data complexity classifier with the data context as the input. Here we assess its quality by testing its performances in-domain (with an ensemble of 3 models) on the RACE++ test set. We additionally compare the performance on the standard input with other possible combinations of the input, as shown in Table \ref{tab:test_comp}. As well as accuracy, macro F1 is reported to account for the imbalance in the complexity level classes. The results for the mode class indicate the baseline performance when the mode class is selected for every example in the test set.
All systems significantly outperform the baseline.
Inputting the context alone is sufficient to get an accuracy close to the full input and when extra information is inputted, the gain is marginal.
Hence, compared with question and options, the context carries a substantial proportion of the information to determine the complexity of a question.

\begin{table}[htbp!]
\small
    \centering
    \begin{tabular}{l|cc|cc}
        \toprule
        \multirow{2}{*}{input format} 
        & \multicolumn{2}{c|}{accuracy} & \multicolumn{2}{c}{F1} \\
        & single & ens & single & ens \\
        \midrule  
        mode class & 61.6 & -- & 25.4 & -- \\
        \midrule
        standard  & $\text{84.7}_{\pm{\text{0.5}}}$ & $\text{87.2}$  & $\text{81.7}_{\pm{\text{1.1}}}$ & 83.7 \\
        context  & $\text{84.9}_{\pm{\text{0.3}}}$ & 85.1  & $\text{81.8}_{\pm{\text{0.8}}}$ & 81.7 \\
        context-question  & $\text{84.7}_{\pm{\text{0.7}}}$ & 86.0 &  $\text{81.8}_{\pm{\text{0.6}}}$ & 82.2 \\
        question-option  & $\text{70.2}_{\pm{\text{0.5}}}$ & 71.3 & $\text{67.3}_{\pm{\text{0.7}}}$  & 68.2 \\
        \bottomrule
    \end{tabular}
    \caption{Accuracy of data complexity evaluators on the RACE++ test set.}
    \label{tab:test_comp}
\end{table}

\subsubsection{Additional models}

In Section \ref{sec:results}, we analyse the influence from different components to the output for two specific models: Llama-2 for the multiple choice reading comprehension task, Roberta for the sentiment classification task. Here we show the component influence is not influenced by the specific choice of the model by showing the consistency of the component influence on the same datasets but evaluated by different models: Llama-2, Roberta and Longformer as in Table \ref{tab:test_accsA}.

{
\centering
\begin{table*}[h]
\small
    \centering
    \begin{tabular}{c|c|cc|ccc|cc}
    \toprule
    \multirow{2}{*}{model}  &\multirow{2}{*}{dataset}& \multicolumn{2}{c|}{accuracy} & \multicolumn{5}{c}{influence}\\
     & & original & para & total & question & context  & context-semantic & context-linguistic \\
    \midrule
    \multirow{3}{*}{MCTest} & RoBerta& 95.3 & 93.0 & 0.254 & 0.140 (55.2\%)& 0.114 (44.8\%) & 0.076 (67.0\%)& 0.037 (33.0\%)\\
    & Longformer & 98.3& 91.3 & 0.285 & 0.152 (53.5\%)& 0.133 (46.5\%) & 0.068 (70.6\%) & 0.028 (29.4\%)\\
    & Llama2 & 92.5 & 85.8 & 0.212 & 0.116 (54.7\%) & 0.096 (45.3\%) & 0.068 (70.6\%) & 0.028 (29.4\%) \\
    \midrule
    \multirow{3}{*}{RACE++} & RoBerta & 84.2 & 81.5 & 0.379 & 0.213 (56.3\%)& 0.166 (43.7\%) & 0.135 (81.4\%)&0.031 (18.6\%)\\
    & Longformer & 81.6 & 79.3 & 0.390& 0.228 (58.6\%)& 0.162 (41.1\%) & 0.135 (83.2\%)& 0.027 (16.8\%)\\
    & Llama2 & 86.0& 82.9 & 0.298 &0.161 (56.1\%)&0.131 (43.9\%)&0.108 (82.5\%)& 0.023 (17.5\%)\\
    \midrule
    \multirow{3}{*}{CMCQRD} & Roberta & 73.5 & 69.4 & 0.383 & 0.287 (74.9\%) & 0.096 (25.1\%) & 0.074 (77.5\%)& 0.022 (22.4\%)\\
    & Longformer & 71.9& 69.8 & 0.467 & 0.326 (69.9\%)& 0.141 (30.1\%) & 0.114 (81.0\%)& 0.027 (19.0\%)\\
    & Llama2 & 79.9 & 69.4 & 0.290 & 0.211 (72.7\%) & 0.079 (27.3\%) & 0.067 (83.8\%) & 0.012 (16.2\%) \\
    \bottomrule
    \end{tabular}
    \caption{Decomposition of total input influence for different models in various multiple-choice reading comprehension datasets.}
    \label{tab:test_accsA}
\end{table*}
}

\subsubsection{Further analysis}

\begin{figure*}[t]
    \centering
    \begin{subfigure}[t]{1\columnwidth}
        \centering
        \includegraphics[width=3in]{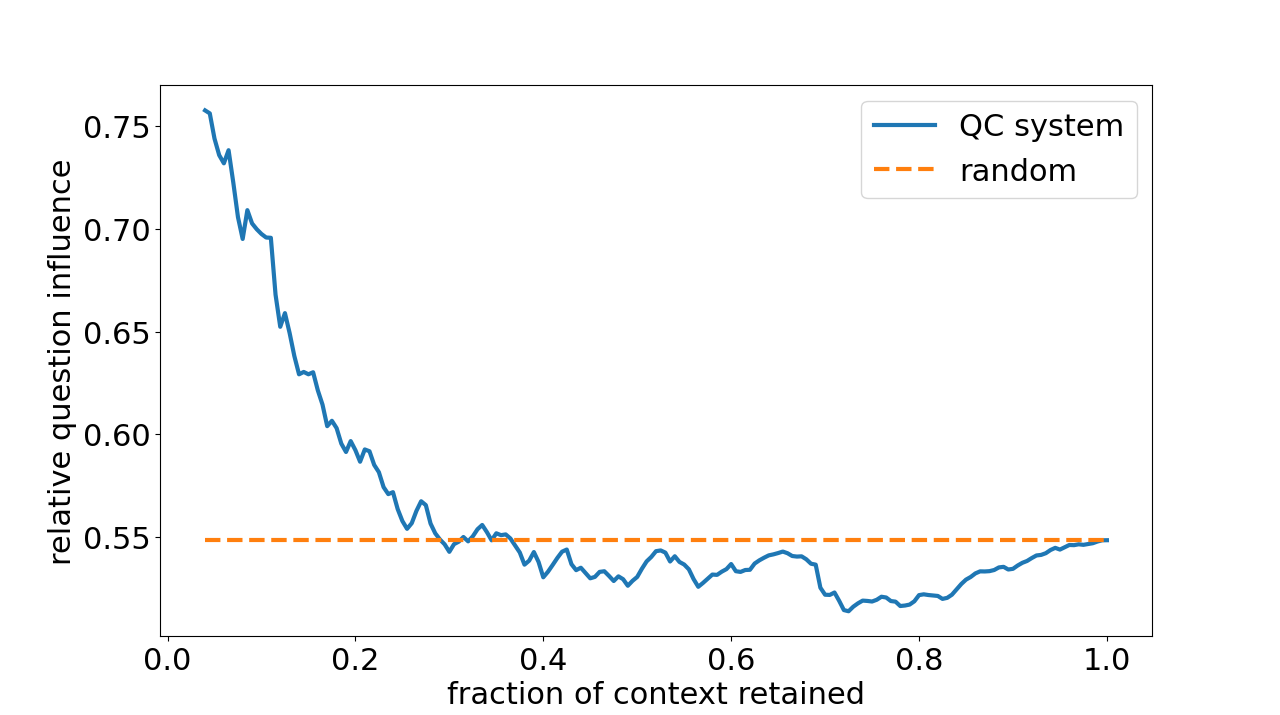}
        \caption{num of words}
        \label{fig:sweep_words}
    \end{subfigure}%
    ~ 
    \begin{subfigure}[t]{1\columnwidth}
        \centering
        \includegraphics[width=3in]{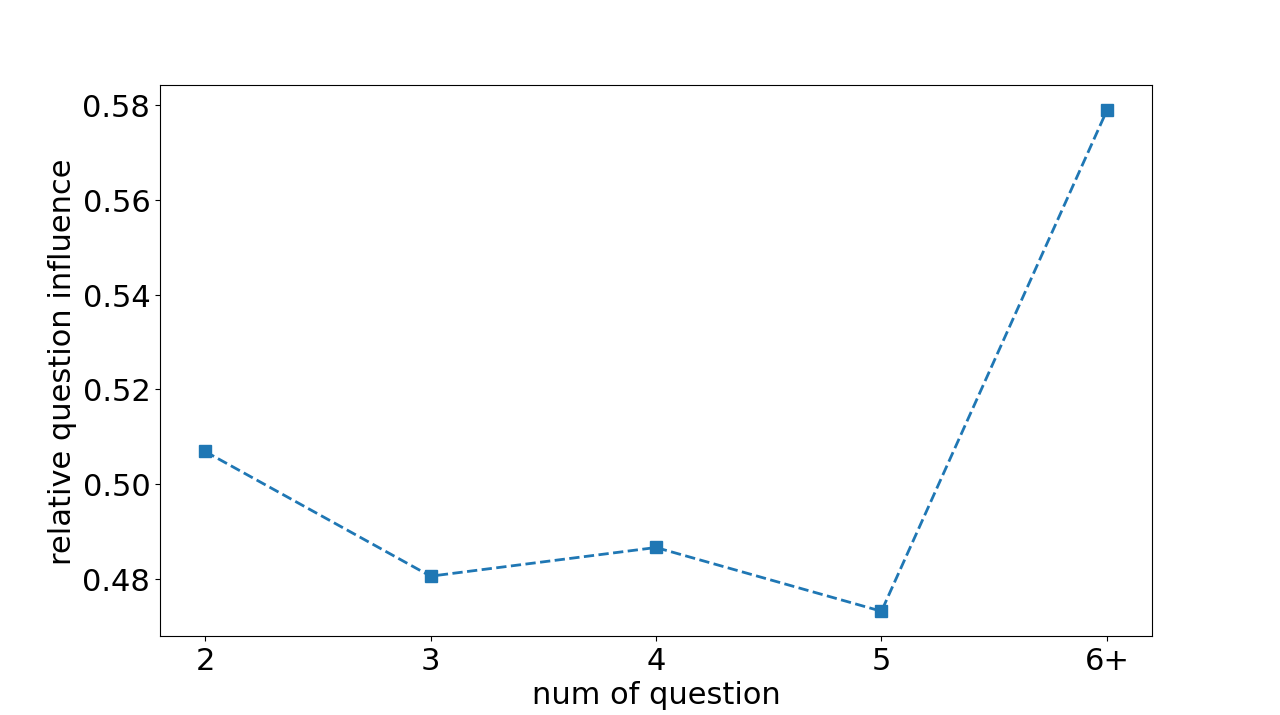}
        \caption{num of questions}
        \label{fig:sweep_questions}
    \end{subfigure}%
    \caption{The relative question influence for a subset of contexts swept in order of length (left) or average number of questions per context (right) for all MCRC datasets with Llama-2.}
    \label{fig:q_inf_sweep}
\end{figure*}

Here we explore other potential factors influencing the relative question influences. From Table \ref{tab:test_data}, there are two marked differences between the datasets: the number of words (length) and the number of questions. To find their influence in the relative question influence score, Figure \ref{fig:sweep_words} and Figure \ref{fig:sweep_questions} show the relative question influence of the subset chosen from the contexts ranked by their number of words or the number of questions of the corresponding context. A strong positive trend between the relative question influence scores and the context length is observed as expected: a longer context naturally has a larger question generation capacity. As shown in Figure \ref{fig:sweep_questions}, the number of questions does not have a direct impact, indicating the results are not affected by the specific number of questions per context.

\subsubsection{Ordering}


For humans taking multiple-choice tests, the role of the context compared to the question may be influenced by the ordering in which they read each of these elements. Similarly, a reading comprehension system may be susceptible to the ordering of the context and the question. Here we compare the influence of the ordering by reversing the standard context followed by question at the input to the question followed by the context .Table \ref{tab:test_accs_ab} demonstrates that the ordering for the automated system does not lead to differing influences on each element. The results here are provided for the finetuned Llama-2 model from Section \ref{sec:model}.

{
\centering
\begin{table*}[htbp!]
\small
    \centering
    \begin{tabular}{c|c|ccc|cc}
    \toprule
    \multirow{2}{*}{dataset} & \multirow{2}{*}{direction} & \multicolumn{5}{c}{influence}\\
     & & total & question & context &  context-semantic & context-linguistic\\
    \midrule
     \multirow{2}{*}{RACE++} & Forward & 0.304  & 0.171 (56.3\%) & 0.133 (43.7\%) & 0.109 (82.1\%) & 0.024 (17.9\%)\\
     & Reverse & 0.305  & 0.172 (56.7\%) & 0.133 (43.6\%) & 0.109 (82.3\%)& 0.024 (17.7\%)\\
     \midrule
     \multirow{2}{*}{MCTest} & Forward & 0.212  & 0.116 (54.7\%) & 0.096 (45.3\%) & 0.068 (70.6\%)& 0.028 (29.4\%)\\
     & Reverse & 0.229  & 0.129 (56.6\%) & 0.100 (43.4\%) & 0.067 (67.2\%)& 0.032 (32.3\%)\\
     \midrule
     \multirow{2}{*}{CMCQRD} & Forward & 0.290  & 0.211 (72.7\%) & 0.079 (27.3\%) & 0.067 (83.8\%)& 0.012 (16.2\%)\\
     & Reverse & 0.278  & 0.204 (73.4\%) & 0.074 (26.6\%) & 0.061 (82.5\%)& 0.013 (17.5\%)\\
    \bottomrule
    \end{tabular}
    \caption{Decomposition of total input influence for different models in various datasets for context-question (forward) vs question-context (reverse) using Llama-2.}
    \label{tab:test_accs_ab}
\end{table*}
}

\subsection{Sentiment classification}
\label{app:res-sc}
For the sentiment classification task, to show the consistency of the element influence among different models, Table \ref{tab:test_sc_all} presents additional results using the BERT model.

\begin{table*}
\small
    \centering
    \begin{tabular}{c|c|cc|ccc}
    \toprule
    \multirow{2}{*}{dataset} & \multirow{2}{*}{model}  & \multicolumn{2}{c|}{accuracy} & \multicolumn{3}{c}{influence}\\
     & & original & para & context  & context-semantic & context-linguistic \\
        \midrule
        \multirow{2}{*}{IMDb}& Roberta & 94.8 & 94.0 & 0.472 & 0.444 (94.2\%) & 0.028 (5.8\%)\\
        & Bert & 93.3 & 92.9 & 0.483 & 0.458 (94.7\%) & 0.025 (5.3\%)\\
        \midrule
        \multirow{2}{*}{Yelp}& Roberta & 94.3 & 93.9 & 0.472 &  0.445 (94.2\%)&  0.027 (5.8\%)\\
        & Bert & 92.9 & 92.6 & 0.518 &  0.488 (94.2\%)&  0.030 (5.8\%)\\
        \midrule
        \multirow{2}{*}{Amazon}& Roberta &  91.0&  89.5 & 0.361 & 0.325 (90.0\%)& 0.036 (10.0\%)\\
        & Bert &  91.2&  90.3 & 0.425 & 0.389 (91.5\%)& 0.036 (8.5\%)\\
        \midrule
        \multirow{2}{*}{Sst2}& Roberta & 87.4 & 82.5 & 0.210 & 0.171 (81.4\%) & 0.039 (18.6\%)\\
        &  Bert & 89.0 & 84.7 & 0.274 & 0.229 (83.5\%) & 0.045 (16.5\%)\\
        \midrule
        \multirow{2}{*}{Tweeteval}& Roberta & 85.2 & 74.5 & 0.570 & 0.469 (82.2\%) & 0.101 (17.8\%)\\
        & Bert & 77.7 & 75.3 & 0.592 & 0.506 (85.5\%) & 0.086 (14.5\%)\\
        \bottomrule
    \end{tabular}
    \caption{Decomposition of total input influence for different models in various sentiment classification dataset}
    \label{tab:test_sc_all}
\end{table*}

\end{document}